\documentclass{article}
\pdfoutput=1

\PassOptionsToPackage{numbers,sort&compress}{natbib}

\usepackage[final]{neurips_2020}

\usepackage[utf8]{inputenc}
\usepackage[T1]{fontenc}
\usepackage{microtype}
\usepackage{textcomp}
\usepackage{graphicx}
\usepackage{xspace}
\usepackage{xcolor}
\usepackage{caption}
\usepackage{booktabs}
\usepackage{amsmath,amsfonts}
\usepackage{makecell}
\usepackage{colortbl}
\usepackage{wrapfig}
\usepackage[subtle]{savetrees}
\usepackage{enumitem}
\setitemize{noitemsep,topsep=2pt,parsep=2pt,partopsep=2pt}

\usepackage{titlesec}
\titleformat{\subsection}[runin]{\normalfont\bfseries}{\thesubsection.}{0.5em}{}[.]
\titlespacing{\subsection}{0pt}{0.5em}{1em}
\titlespacing{\paragraph}{0pt}{0.5em}{1em}

\usepackage[breaklinks,colorlinks,urlcolor=blue,bookmarks=false]{hyperref}
\hypersetup{ %
	pdftitle={BlockGAN: Learning 3D Object-aware Scene Representations from Unlabelled Images},
	pdfauthor={Thu Nguyen-Phuoc, Christian Richardt, Long Mai, Yong-Liang Yang, Niloy Mitra},
	pdfsubject={NeurIPS 2020}
}
\usepackage{etoolbox}
\makeatletter
\patchcmd\@combinedblfloats{\box\@outputbox}{\unvbox\@outputbox}{}{%
}%
\makeatother

\newcommand{\abs}[1]{\left\lvert#1\right\rvert}

\newcommand{\name}{BlockGAN\xspace}
\newcommand{\pose}{\boldsymbol{\theta}}

\newcommand{\gline}{\arrayrulecolor{gray}\hline\arrayrulecolor{black}}
\newcommand{\diff}[1]{\textcolor{blue!70}{#1}}

\begin{document}
\defcitealias{blender}{Blender}

\title{\name: Learning 3D Object-aware Scene Representations from Unlabelled Images}

\author{%
	Thu Nguyen-Phuoc \\ University of Bath \And
	Christian Richardt \\ University of Bath \And
	Long Mai \\ Adobe Research \AND
	Yong-Liang Yang \\ University of Bath \And
	Niloy Mitra \\ Adobe Research \& UCL \\
}

\maketitle

\begin{abstract}
We present \name, an image generative model that learns object-aware 3D scene representations directly from unlabelled 2D images. 
Current work on scene representation learning either ignores scene background or treats the whole scene as one object.
Meanwhile, work that considers scene compositionality treats scene objects only as image patches or 2D layers with alpha maps.
Inspired by the computer graphics pipeline, we design \name to learn to first generate 3D features of background and foreground objects, then combine them into 3D features for the whole scene, and finally render them into realistic images.
This allows \name to reason over occlusion and interaction between objects' appearance, such as shadow and lighting, and provides control over each object's 3D pose and identity, while maintaining image realism.
\name is trained end-to-end, using only unlabelled single images, without the need for 3D geometry, pose labels, object masks, or multiple views of the same scene.
Our experiments show that using explicit 3D features to represent objects allows \name to learn disentangled representations both in terms of objects (foreground and background) and their properties (pose and identity).
Our code is available at \href{https://www.github.com/thunguyenphuoc/BlockGAN}{https://github.com/thunguyenphuoc/BlockGAN}.
\end{abstract}

\section{Introduction}

The computer graphics pipeline has achieved impressive results in generating high-quality images, while offering users a great level of freedom and controllability over the generated images.
This has many applications in creating and editing content for the creative industries, such as films, games, scientific visualisation, and more recently, in generating training data for computer vision tasks.
However, the current pipeline, ranging from generating 3D geometry and textures, rendering, compositing and image post-processing, can be very expensive in terms of labour, time, and costs.

Recent image generative models, in particular generative adversarial networks \citep[GANs;][]{NIPS2014_5423}, have greatly improved the visual fidelity and resolution of generated images \cite{karras2018, karras2019, brock2018large}.
Conditional GANs \cite{mirza2014conditional} allow users to manipulate images, but require labels during training.
Recent work on unsupervised disentangled representations using GANs \cite{Chen2016, karras2019, Nguyen-Phuoc_2019_ICCV} relaxes this need for labels.
The ability to produce high-quality, controllable images has made GANs an increasingly attractive alternative to the traditional graphics pipeline for content generation.
However, most work focuses on \emph{property} disentanglement, such as shape, pose and appearance, without considering the compositionality of the images, i.e., scenes being made up of multiple objects.
Therefore, they do not offer control over individual objects in a way that respects the interaction of objects, such as consistent lighting and shadows.
This is a major limitation of current image generative models, compared to the graphics pipeline, where 3D objects are modelled individually in terms of geometry and appearance, and combined into 3D scenes with consistent lighting.

Even when considering object compositionality, most approaches treat objects as 2D layers combined using alpha compositing \cite{yang_2017lr, GAN_composition, engelcke2020genesis}.
Moreover, they also assume that each object's appearance is independent \cite{bielski2019emergence,NIPS2019_9434, engelcke2020genesis}.
While this layering approach has led to good results in terms of object separation and visual fidelity, it is fundamentally limited by the choice of 2D representation.
Firstly, it is hard to manipulate properties that require 3D understanding, such as pose or perspective.
Secondly, object layers tend to bake in appearance and cannot adequately represent view-specific appearance, such as shadows or material highlights changing as objects move around in the scene.
Finally, it is non-trivial to model the appearance interactions between objects, such as scene lighting that affects objects' shadows on a background.

We introduce \name, a generative adversarial network that learns 3D object-oriented scene representations directly from unlabelled 2D images.
Instead of learning 2D layers of objects and combining them with alpha compositing, \name learns to generate 3D object features and to combine them into deep 3D scene features that are projected and rendered as 2D images.
This process closely resembles the computer graphics pipeline where scenes are modelled in 3D, enabling reasoning over occlusion and interaction between object appearance, such as shadows or highlights.
During test time, each object's pose can be manipulated using 3D transforms directly applied to the object's deep 3D features.
We can also add new objects and remove existing objects in the generated image by changing the number of 3D object features in the 3D scene features at inference time.
This shows that \name has learnt a non-trivial representation of objects and their interaction, instead of merely memorizing images.

\name is trained end-to-end in an unsupervised manner directly from unlabelled 2D images, without any multi-view images, paired images, pose labels, or 3D shapes.
We experiment with \name on a variety of synthetic and natural image datasets.
In summary, our main contributions are: 
\begin{itemize}
  \item \name, an unsupervised image generative model that learns an object-aware 3D scene representation directly from unlabelled 2D images, disentangling both between objects and individual object properties (pose and identity);
  
  \item showing that \name can learn to separate objects even from cluttered backgrounds; and 
  
  \item demonstrating that \name's object features can be added, removed and manipulated to create novel scenes that are not observed during training.
\end{itemize}

\section{Related work}

\paragraph{GANs.}

Unsupervised GANs %
learn to map samples from a latent distribution to data categorised as real by a discriminator network.
Conditional GANs %
enable control over the generated image content, but require labels during training.
Recent work on unsupervised disentangled representation learning using GANs provides controllability over the final images without the need for labels.
Loss functions can be designed to maximize mutual information between generated images and latent variables \cite{Chen2016, jeon2019ibgan}.
However, these models do not guarantee which factors can be learnt, and have limited success when applied to natural images.
Network architectures can play a vital role in both improving training stability \cite{chen2018on} and controllability of generated images \cite{karras2019, Nguyen-Phuoc_2019_ICCV}.
We also focus on designing an appropriate architecture to learn object-level disentangled representations.
We show that injecting inductive biases about how the 3D world is composed of 3D objects enables \name to learn 3D object-aware scene representations directly from 2D images, thus providing control over both 3D pose and appearance of individual objects.

\paragraph{3D-aware neural image synthesis.}

Introducing 3D structures into neural networks can improve the quality \cite{tvsn_cvpr2017, NIPS2018_8014, Rhodin2018, Sitzmann_2019} and controllability of the image generation process \cite{Nguyen-Phuoc_2019_ICCV, olszewski2019tbn, NIPS2018_7297}.
This can be achieved with explicit 3D representations, like appearance flow \cite{ZhouTSME2016}, occupancy voxel grids \cite{NIPS2018_7297,rematas2019neural}, meshes, or shape templates \cite{Kossaifi2018, Shu_2018_ECCV, YaoHZWTFT2018}, in conjunction with handcrafted differentiable renderers \cite{Loper2014, Henzler_2019_ICCV, NIPS2019_9156, Liu_2019_ICCV}.
Renderable deep 3D representations can also be learnt directly from images \cite{Nguyen-Phuoc_2019_ICCV, Sitzmann2018, Sitzmann_2019}.
HoloGAN \cite{Nguyen-Phuoc_2019_ICCV} further shows that adding inductive biases about the 3D structure of the world enables unsupervised disentangled feature learning between shape, appearance and pose.
However, these learnt representations are either object-centric (i.e., no background), or treat the whole scene as one object.
Thus, they do not consider scene compositionality, i.e., components that can move independently.
\name, in contrast, is designed to learn object-aware 3D representations that are combined into a unified 3D scene representation.

\paragraph{Object-aware image synthesis.}

Recent methods decompose image synthesis into generating components like layers or image patches, and combining them into the final image \cite{yang_2017lr, Kwak2016, GAN_composition}.
This includes conditional GANs that use segmentation masks \cite{turkoglu2019layer, papadopoulos19cvpr}, scene graphs \cite{johnson2018image}, object labels, key points or bounding boxes \cite{hinz2019generating, NIPS2016_6111}, which have shown impressive results for natural image datasets.
Recently, unsupervised methods \cite{NIPS2016_6230, NIPS2018_8079, GAN_composition, engelcke2020genesis, anciukevicius20corr, Yang2020ManipulateObjs} learned object disentanglement for multi-object scenes on simpler synthetic datasets (single-colour objects, simple lighting, and material).
Other approaches successfully separate foreground from background objects in natural images, but make strong assumptions about the size of objects \cite{yang_2017lr} or independent object appearance \cite{bielski2019emergence,NIPS2019_9434}.
These methods treat object components as image patches or 2D layers with corresponding masks, which are combined via alpha compositing at the pixel level to generate the final stage.
The work closest to ours learns to generate multiple 3D primitives (cuboids, spheres and point clouds), renders them into \emph{separate} 2D layers with a handcrafted differentiable renderer, and alpha-composes them based on their depth ordering to create the final image \citep{LiaoSMG2020}.
Despite the explicit 3D geometry, this method does not handle cluttered backgrounds and requires extra supervision in the shape of labelled images with and without foreground objects.

\name takes a different approach.
We treat objects as \emph{learnt 3D features} with corresponding 3D poses, and learn to combine them into 3D scene features.
Not only does this provide control over 3D pose, but also enables learning of realistic lighting and shadows.
Our approach allows adding more objects into the 3D scene features to generate images with multiple objects, which are not observed at training time.

\begin{figure*}
	\includegraphics[width=\linewidth]{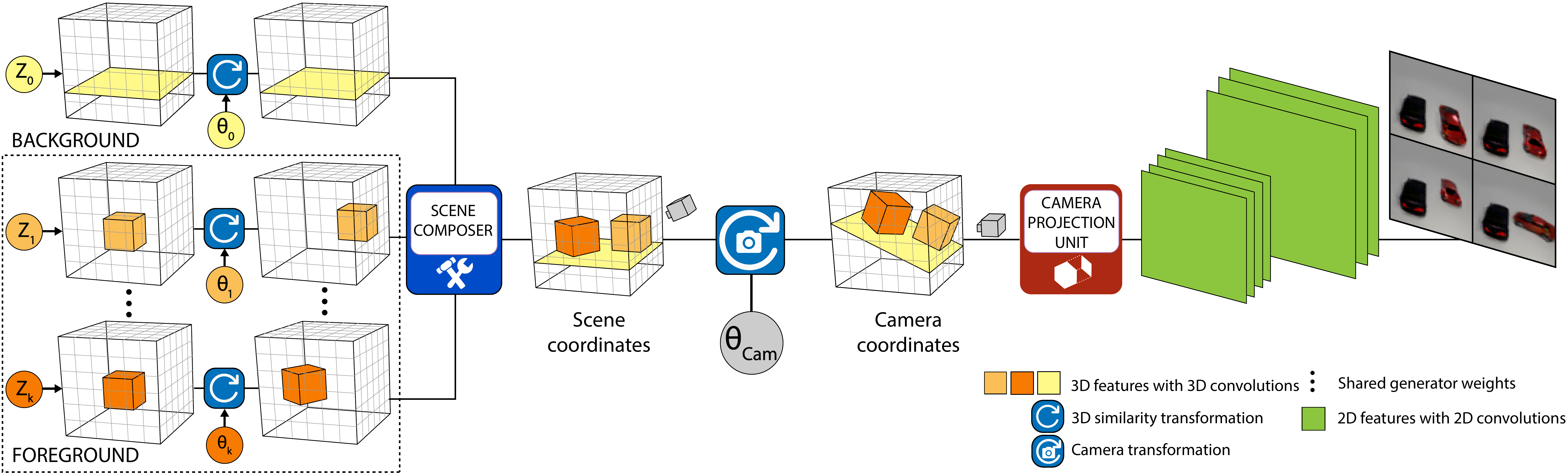}%
	\caption{\label{fig:ArchitectureDiagram}
		\name's generator network.
		Each noise vector $\mathbf{z}_i$ is mapped to deep 3D object features, which are transformed to the desired 3D pose $\pose_i$. %
		Object features are combined into 3D scene features, where the camera pose $\pose_\text{cam}$ is applied, before projection to 2D features that produce the image~$\mathbf{x}$.
	}
\end{figure*}

\section{Method}

Inspired by the computer graphics pipeline, we assume that each image $\mathbf{x}$ is a rendered 2D image of a 3D scene composed of $K$ 3D foreground objects $\{O_1, \ldots, O_K\}$ in addition to the background $O_0$:
\begin{equation}
\label{eq:}
    \mathbf{x} = p\big(f(
        \underbrace{O_0,}_\text{background}
        \underbrace{O_1, \;\ldots,\; O_K}_\text{foreground}
    )\big) \text{,}
\end{equation}
where the function $f$ combines multiple objects into unified scene features that are projected to the image $\mathbf{x}$ by $p$.
We assume each object $O_i$ is defined in a canonical orientation and generated from a noise vector $\mathbf{z}_i$ by a function $g_i$ before being individually posed using parameters $\pose_i$: $O_i = g_i(\mathbf{z}_i, \pose_i)$.

We inject the inductive bias of compositionality of the 3D world into \name in two ways.
(1) The generator is designed to first generate 3D features for each object independently, before transforming and combining them into unified scene features, in which objects interact.
(2) Unlike other methods that use 2D image patches or layers to represent objects, \name directly learns from unlabelled images how to generate objects as 3D features.
This allows our model to disentangle the scene into separate 3D objects and allows the generator to reason over 3D space, enabling object pose manipulation and appearance interaction between objects.
\name, therefore, learns to both generate and render the scene features into images that can fool the discriminator.

Figure \ref{fig:ArchitectureDiagram} illustrates the \name generator architecture.
Each noise vector $\mathbf{z}_i$ is mapped to 3D object features~$O_i$. %
Objects are then transformed according to their pose $\pose_i$ using a 3D similarity transform, before being combined into 3D scene features using the \emph{scene composer} $f$. %
The scene features are transformed into the camera coordinate system before being projected to 2D features to render the final images using the \emph{camera projector} function $p$. %
During training, we randomly sample both the noise vectors $\mathbf{z}_i$ and poses~$\pose_i$.
During test time, objects can be generated with a given identity $\mathbf{z}_i$ in the desired pose $\pose_i$.

\name is trained end-to-end using only unlabelled 2D images, without the need for any labels, such as poses, 3D shapes, multi-view inputs, masks, or geometry priors like shape templates, symmetry or smoothness terms. %
We next explain each component of the generator in more detail.

\subsection{Learning 3D object representations}
\label{sec:object feature}

Each object $O_i \!\in\! \mathbb{R}^{H_o \times W_o \times D_o \times C_o}$ is a deep 3D feature grid generated by $O_i = g_i(\mathbf{z}_i, \pose_i)$, where $g_i$ is an object generator that takes as input a noise vector $\mathbf{z}_i$ controlling the object appearance, and the object's 3D pose $\pose_i = (s_i, \mathbf{R}_i, \mathbf{t}_i)$, which comprises its uniform scale $s_i \!\in\! \mathbb{R}$, rotation $\mathbf{R}_i \!\in\! \mathrm{SE}(3)$ and translation $\mathbf{t}_i \!\in\! \mathbb{R}^3$.
The object generator $g_i$ is specific to each category of objects, and is shared between objects of the same category.
We assume that 3D scenes consist of at least two objects: the background $O_0$ and one or more foreground objects $\{O_1, \ldots, O_K\}$.
This is different to object-centric methods that only assume a single object with a simple white background \cite{Sitzmann2018}, or only deal with static scenes whose object components cannot move independently \citep{Nguyen-Phuoc_2019_ICCV}.
We show that, even when \name is trained with only one foreground and background object, we can add an arbitrary number of foreground objects to the scene at test time.

\begin{wrapfigure}[13]{r}{0.4\linewidth}
	\vspace{-12pt}
	\includegraphics[width=\linewidth,trim=0 10 265 20,clip]{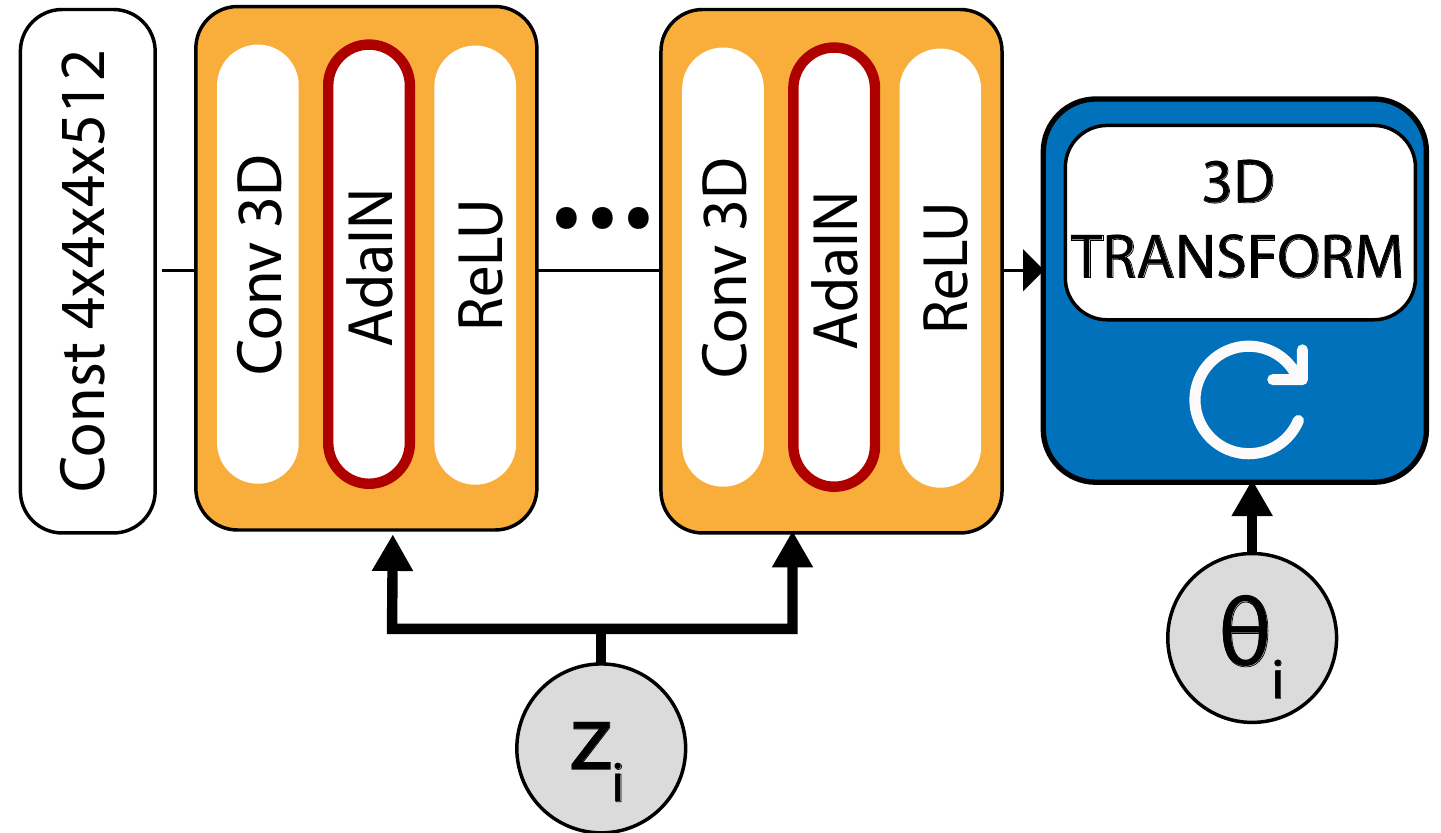}%
\caption{\label{fig:ObjectGenerator}
	\name's object generator. %
	Each object starts with a constant tensor that is learnt with the rest of the network.
}
\end{wrapfigure}
To generate 3D object features, \name implements the style-based strategy, which helps to disentangle between pose and identity \cite{Nguyen-Phuoc_2019_ICCV} while improving training stability \cite{karras2019}.
As illustrated in Figure~\ref{fig:ObjectGenerator}, the noise vector $\mathbf{z}_i$ is mapped to affine parameters – the ``style controller'' – for adaptive instance normalization \citep[AdaIN;][]{huang2017adain} after each 3D convolution layer. 
However, unlike HoloGAN \cite{Nguyen-Phuoc_2019_ICCV}, which learns 3D features directly for the whole scene, \name learns 3D features for \emph{each} object, which are then transformed to their target poses using similarity transforms, and combined into 3D \emph{scene} features.
We implement these 3D similarity transforms by trilinear resampling of the 3D features according to the translation, rotation and scale parameters $\pose_i$;
samples falling outside the feature tensor are clamped to zero.
This allows \name to not only separate object pose from identity, but also to disentangle multiple objects in the same scene.

\subsection{Scene composer function}
\label{sec:scene composer}

We combine the 3D object features $\{O_i\}$ into scene features
$    S = f(O_0,\; O_1,\; \ldots,\; O_K) \in \mathbb{R}^{H_s \times W_s \times D_s \times C_s}$
using a scene composer function $f$.
For this, we use the element-wise maximum as it achieves the best image quality compared to element-wise summation and a multi-layer perceptron (MLP); please see our supplemental document for an ablation.
Additionally, the maximum is invariant to permutation and allows a flexible number of input objects to add new objects into the scene features during test time, even when trained with fewer objects (see Section~\ref{sec:manipulaton}).

\subsection{Learning to render}
\label{sec:render}

Instead of using a handcrafted differentiable renderer, we aim to learn rendering directly from unlabelled images.
HoloGAN showed that this approach is more expressive as it is capable of handling unlabelled, natural image data.
However, their projection model is limited to a weak perspective, which does not support foreshortening – an effect that is observed when objects are close to real (perspective) cameras.
We therefore introduce a graphics-based perspective projection function $p \colon \mathbb{R}^{H_s \times W_s \times D_s \times C_s} \mapsto \mathbb{R}^{H_c \times W_c \times C_c}$ that transforms the 3D scene features into camera space using a projective transform, and then learns the projection of the 3D features to a 2D feature map.

\begin{wrapfigure}[14]{r}{0.5\linewidth}
	\vspace{-13pt}
	\includegraphics[width=\linewidth]{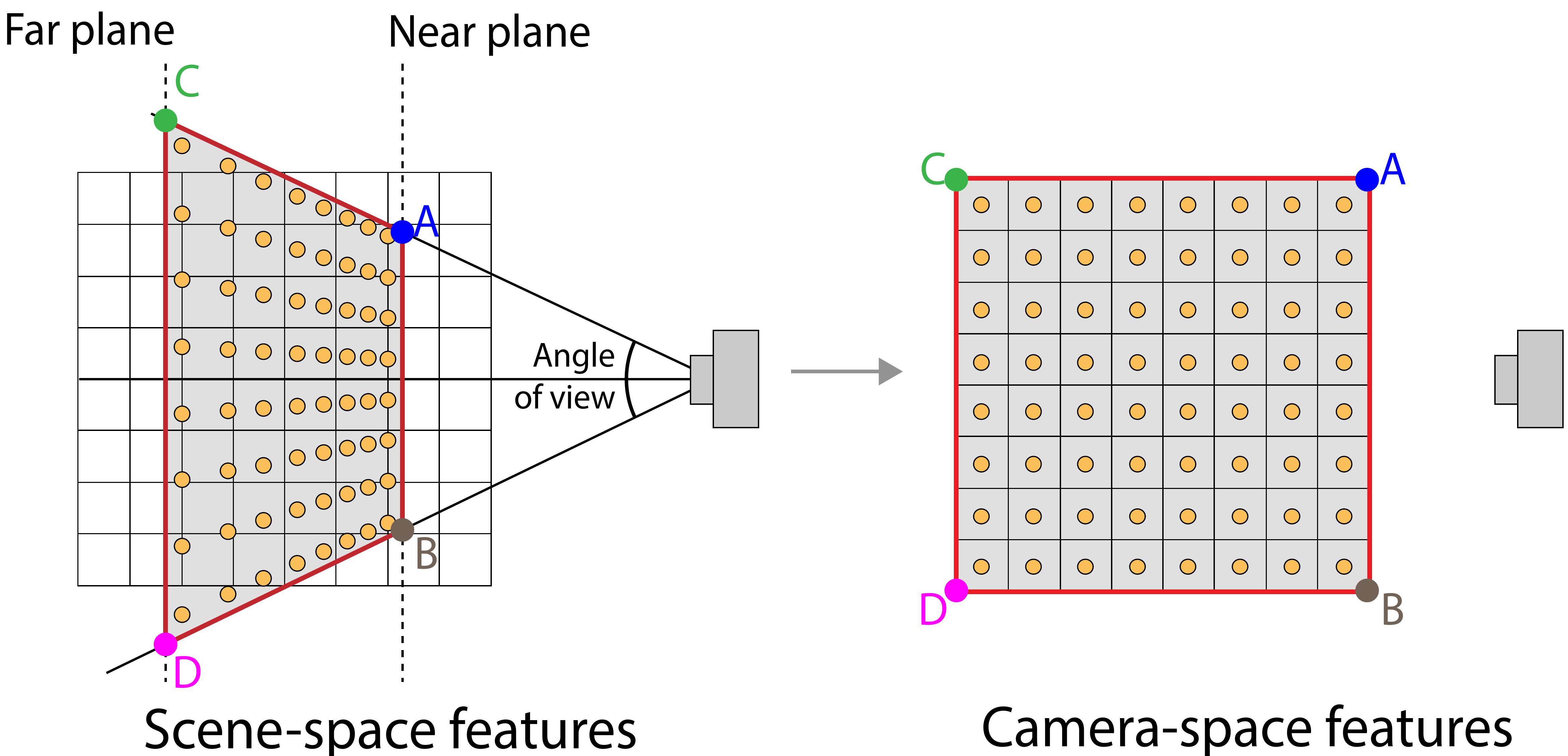}%
	\caption{\label{fig:camera-resampling}
		\textbf{Left:}
		The camera's viewing volume (frustum) overlaid on scene-space features.
		We trilinearly resample the scene features based on the viewing volume at the orange dots.
		\textbf{Right:}
		The resulting camera-space features before projection to 2D.
	}
\end{wrapfigure}
The computer graphics pipeline implements perspective projection using a projective transform that converts objects from world coordinates (our scene space) to camera coordinates \cite{MarscSAGHJMRTWW2015}.
We implement this camera transform like the similarity transforms used to manipulate objects in Section~\ref{sec:object feature}, by resampling the 3D scene features according to the viewing volume (frustum) of the virtual perspective camera (see Figure~\ref{fig:camera-resampling}).
For correct perspective projection, this transform must be a projective transform, the superset of similarity transforms \cite{Yan2016}.
Specifically, the viewing frustum, in scene space, can be defined relative to the camera's pose $\pose_\text{cam}$ using the angle of view, and the distance of the near and far planes.
The camera-space features are a new 3D tensor of features, of size $H_c \!\times\! W_c \!\times\! D_c \!\times\! C_s$, whose corners are mapped to the corners of the camera's viewing frustum using the unique projective 3D transform computed from the coordinates of corresponding corners using the direct linear transform \citep{HartlZ2004}.

In practice, we combine the object and camera transforms into a single transform by multiplying both transform matrices and resampling the object features in a single step, directly from object to camera space.
This is computationally more efficient than resampling twice, and advantageous from a sampling theory point of view, as the features are only interpolated once, not twice, and thus less information is lost by the resampling.
The combined transform is a fixed, differentiable function with parameters $(\pose_i, \pose_\text{cam})$.
The individual objects are then combined in camera space before the final projection.

After the camera transform, the 3D features are projected into view-specific 2D feature maps using the \emph{learnt} camera projection $p' \colon \mathbb{R}^{H_c \!\times\! W_c \!\times\! D_c \!\times\! C_s} \!\mapsto\! \mathbb{R}^{H_c \!\times\! W_c \!\times\! C_c}$.
This function ensures that occlusion correctly shows nearby objects in front of distant objects.
Following the RenderNet projection unit \citep{NIPS2018_8014},
we reshape the 3D camera-space features (with depth $D_c$ and $C_s$ channels) into a 2D feature map with $(D_c \!\cdot\! C_s)$ channels, followed by a per-pixel MLP (i.e., 1$\times$1 convolution) that outputs $C_c$ channels.
We choose to use this learnt renderer following HoloGAN, which shows the effectiveness of the renderer in learning powerful 3D representations directly from unlabelled images.
This is different from the supervised multi-view setting with pose labels in the renderer of DeepVoxels \cite{Sitzmann2018}, which learns occlusion values, or Neural Volumes \cite{Lombardi2019} and NeRF \cite{ MildeSTBRN2020}, which learn explicit density values.

\subsection{Loss functions}
\label{sec:loss}

We train \name adversarially using the non-saturating GAN loss \cite{NIPS2014_5423}. 
For natural images with cluttered backgrounds, we also add a style discriminator loss \citep{Nguyen-Phuoc_2019_ICCV}.
In addition to classifying the images as real or fake, this discriminator also looks at images at the feature level.
Given image features $\mathbf{\Phi}_l$ at layer $l$, the style discriminator classifies the mean $\boldsymbol{\mu}(\mathbf{\Phi}_l)$ and standard deviation $\boldsymbol{\sigma}(\mathbf{\Phi}_l)$ over the spatial dimensions, which describe the image ``style'' \cite{huang2017adain}.
This more powerful discriminator discourages the foreground generator to include parts of the background within the foreground object(s).
We provide detailed network and loss definitions in the supplemental material.

\section{Experiments}

\begin{figure*}
  	\centering
	\includegraphics[width=\linewidth]{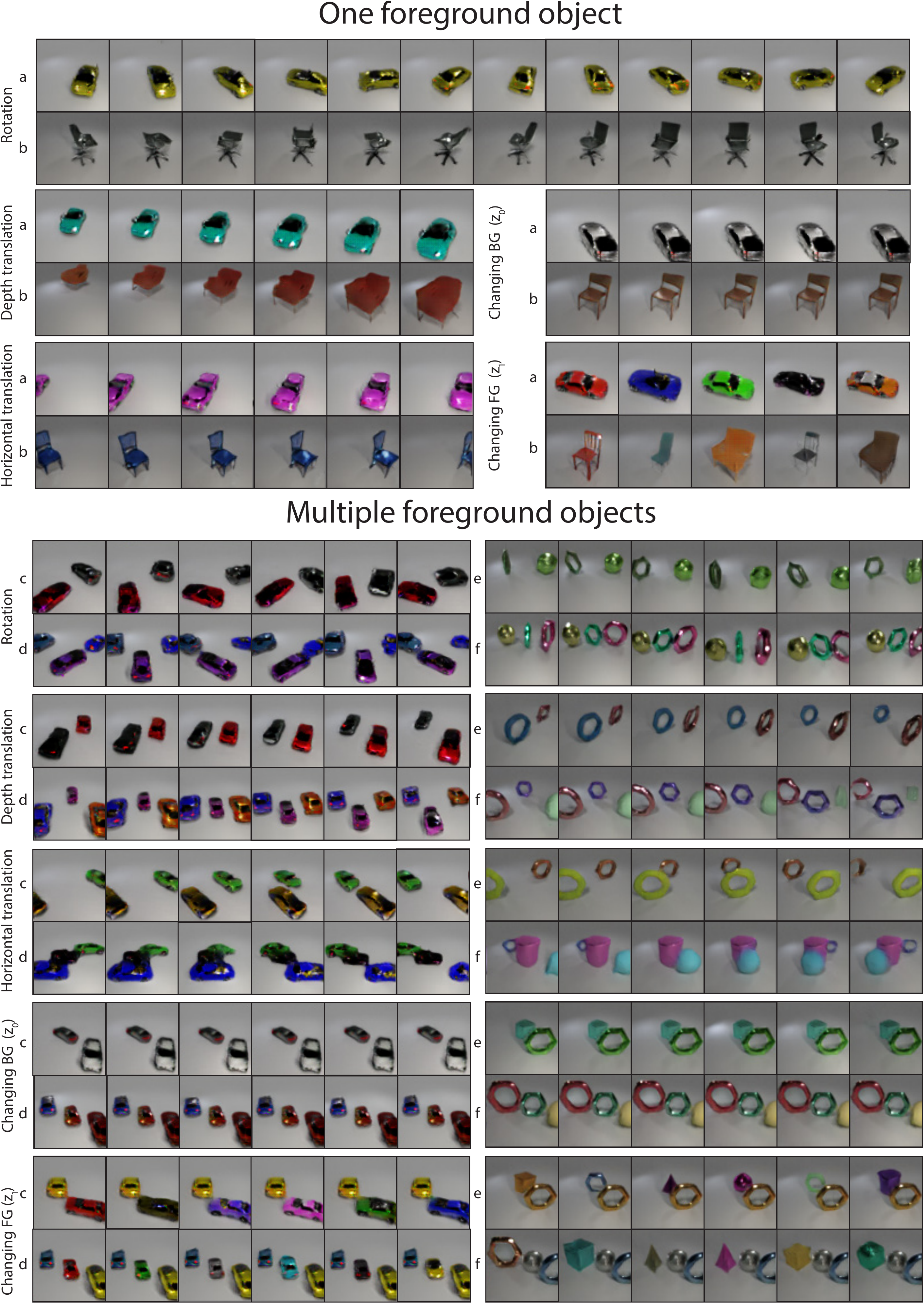}%
	\caption{\label{fig:objectManipulate}
		 \name enables explicit spatial manipulation of individual objects (rotation, translation) and changing the identity of background or foreground objects across different datasets:
		 (a)~\textsc{Synth-Car1}, 
		 (b)~\textsc{Synth-Chair1}, 
		 (c)~\textsc{Synth-Car2},
		 (d)~\textsc{Synth-Car3},
		 (e)~\textsc{CLEVR2} and
		 (f)~\textsc{CLEVR3}.
		 Notice how the shadows and highlights change as objects move around in the scene, and how changing the background lighting affects the appearance of foreground objects.
		 Figure~\ref{fig:linear_interp} shows similar results on natural images. 
		 Please refer to the supplemental material for animated results.
	}
\end{figure*}

\paragraph{Datasets.}

We train \name on images at 64$\times$64\ pixels, with increasing complexity in terms of number of foreground objects (1–4) and texture (synthetic images with simple shapes and simple to natural images with complex texture and cluttered background).
These datasets include the synthetic CLEVR$n$ \cite{johnson2017clevr}, \textsc{Synth-Car$n$} and \textsc{Synth-Chair$n$}, and the real \textsc{Real-Car} \cite{Yangg2015}, where $n$ is the number of foreground objects.
Additional details and results are included in the supplementary material.
 
\paragraph{Implementation details.}

We assume a fixed and known number of objects
of the same type.
Fore- and background generators have similar architectures and the same number of output channels, but foreground generators have twice as many channels in the learnt constant tensor.
Since foreground objects are smaller than the background, we set scale=1 for the background object, and randomly sample scales $<$1 for foreground objects.
Please see our supplemental material for more implementation details and an ablation experiment.
We make our code publicly available at \href{https://www.github.com/thunguyenphuoc/BlockGAN}{github.com/thunguyenphuoc/BlockGAN}.

\subsection{Qualitative results}

Despite being trained with only unlabelled images, Figure~\ref{fig:objectManipulate} shows that \name learns to disentangle different objects within a scene: foreground from background, and between multiple foreground objects.
More importantly, \name also provides explicit control and enables smooth manipulation of each object's pose $\pose_i$ and identity $\mathbf{z}_i$.
Figure~\ref{fig:linear_interp} shows results on natural images with a cluttered background, where \name is still able to separate objects and enables 3D object-centric modifications.
Since \name combines deep object features into scene features, changes in an object's properties also influence its shadows, and highlights adapt to the object's movement.
These effects can be better observed in the supplementary animations.

\subsection{Quantitative results}

We evaluate the visual fidelity of \name's results using Kernel Inception Distance \citep[KID;][]{binkowski2018demystifying}, which has an unbiased estimator and works even for a small number of images.
Note that KID does \emph{not} measure the quality of object disentanglement, which is the main contribution of \name.
We first compare with a vanilla GAN \citep[WGAN-GP;][]{NIPS2017_7159} using a publicly available implementation\footnote{\href{https://github.com/LynnHo/DCGAN-LSGAN-WGAN-WGAN-GP-Tensorflow}{https://github.com/LynnHo/DCGAN-LSGAN-WGAN-WGAN-GP-Tensorflow}}\!\!.
Secondly, we compare with LR-GAN \cite{yang_2017lr}, a 2D-based method that learns to generate image background and foregrounds separately and recursively.
Finally, we compare with HoloGAN, which learns 3D scene representations that separate camera pose and identity, but does not consider object disentanglement.
For LR-GAN and HoloGAN, we use the authors' code.
We tune hyperparameters and then compute the KID for 10,000 images generated by each model (samples by all methods are included in the supplementary material).
Table~\ref{table:KID} shows that \name generates images with competitive or better visual fidelity than other methods.

\begin{table}%
\caption{\label{table:KID}%
	KID estimates (mean $\pm$ std), lower is better, between real images and images generated by \name and other GANs.
	\name achieves competitive KID scores while providing control of each object in the generated images (which is not measured by KID).
	}\vspace{0.5em}
\centering
\resizebox{0.85\linewidth}{!}
{%
\renewcommand*{\arraystretch}{1.2}%
\newcommand{\size}{{\footnotesize 64$\times$64}}%
\begin{tabular}{lcccc}
	\toprule
	\thead[bl]{\textbf{Method}}            & \thead{\textsc{Synth-Car1} \\ \size} & \thead{\textsc{Synth-Chair1}\\\size} & \thead{\textsc{Real-Car}\\\size} & \thead{\textsc{CLEVR2}\\\size} \\ \midrule
	WGAN-GP \citep{NIPS2017_7159}          &          0.141 $\pm$ 0.002           &          0.111 $\pm$ 0.002          &      0.035 $\pm$ 0.001      & 0.076 $\pm$ 0.002              \\
	LR-GAN \citep{yang_2017lr}             &      \textbf{0.038 $\pm$ 0.001}      &          0.036 $\pm$ 0.002          & \textbf{0.014 $\pm$ 0.001}  & 0.052 $\pm$ 0.001              \\
	HoloGAN \citep{Nguyen-Phuoc_2019_ICCV} &          0.070 $\pm$ 0.001           &          0.058 $\pm$ 0.002          &      0.028 $\pm$ 0.002      & 0.032 $\pm$ 0.001             \\
	\name (ours)                           &          0.039 $\pm$ 0.001           &     \textbf{0.031 $\pm$ 0.001}      &      0.016 $\pm$ 0.001      & \textbf{0.021 $\pm$ 0.001}             \\ \bottomrule
\end{tabular}}%
\end{table}

\subsection{Scene manipulation beyond training data}
\label{sec:manipulaton}

\begin{figure}%
	\centering
	\includegraphics[width=\linewidth]{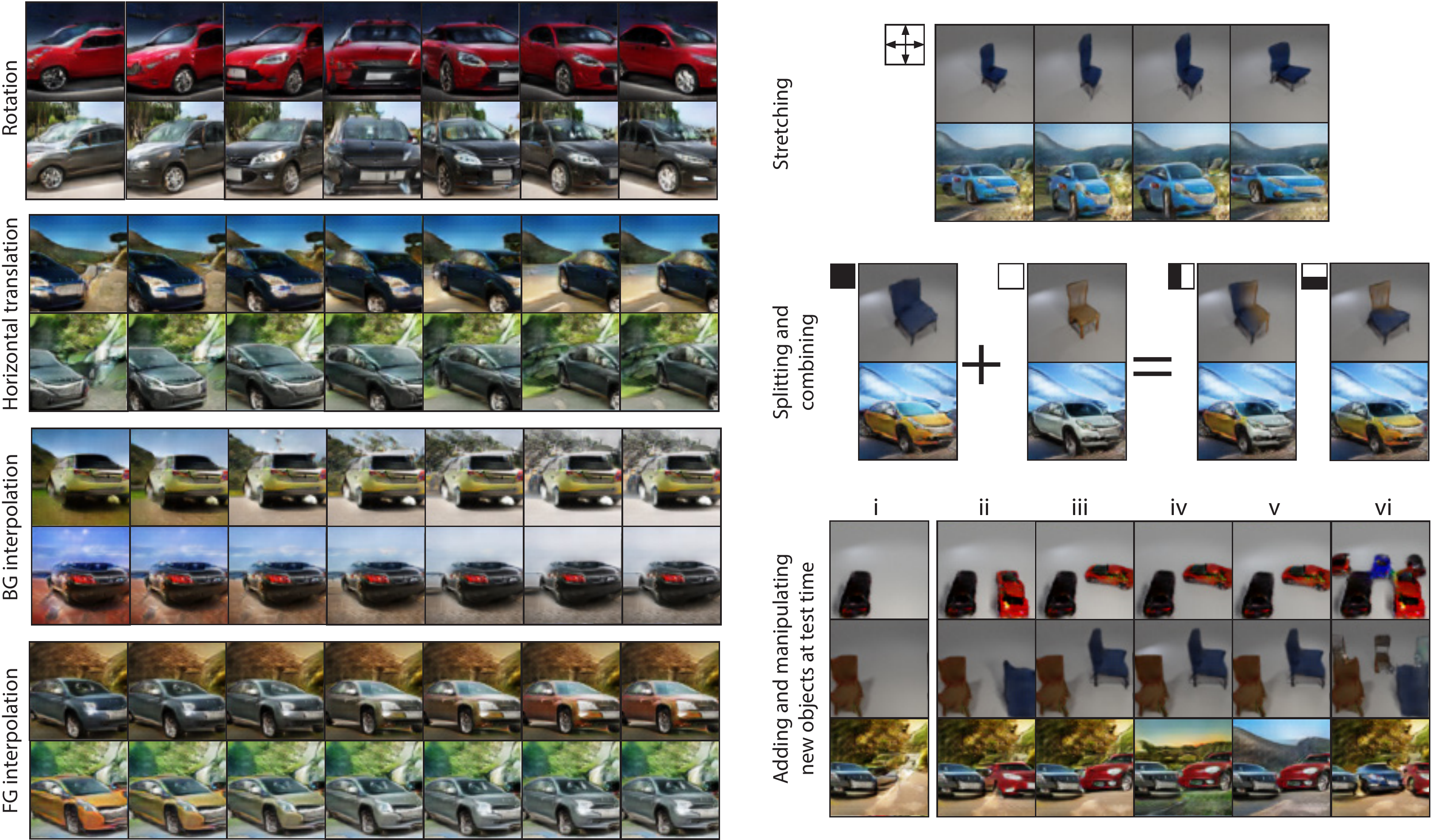}%
	\caption{\label{fig:linear_interp}\label{fig:split_combine}%
		\textbf{Left:} \textsc{Real-Car}.
		Even for natural images with cluttered backgrounds, \name can still disentangle objects in a scene well.
		Note that interpolating the background ($\mathbf{z}_0$) affects the appearance of the car in a meaningful way, showing the benefit of 3D scene features.
		\textbf{Right:}
		Test-time geometric modification of the learnt 3D object features (unless stated, background is fixed): stretching (top), splitting and combining (middle), and adding and manipulating new objects after training (bottom).
		Bottom row shows 
		(i)~Original scene, 
		(ii)~new object added, 
		(iii)~manipulated, 
		(iv,v)~different background appearance, and 
		(vi)~more objects added.
		Note the realistic lighting and shadows.
	    }
\end{figure}

We show that at test time, 3D object features learnt by \name can be realistically manipulated in ways that have not been observed during training time. 
First, we show that the learnt 3D object features can also be reused to add more objects to the scene at test time, thanks to the compositionality inductive bias and our choice of scene composer function.
Firstly, we use \name trained on datasets with only \emph{one} foreground object and \emph{one} background, and show that more foreground objects of the same category can be added to the same scene at \emph{test time}.
Figure~\ref{fig:split_combine} shows that 2–4 new objects are added and manipulated just like the original objects while maintaining
\begin{wrapfigure}[10]{r}{5cm}
	\vspace{-11pt}
	\includegraphics[width=\linewidth]{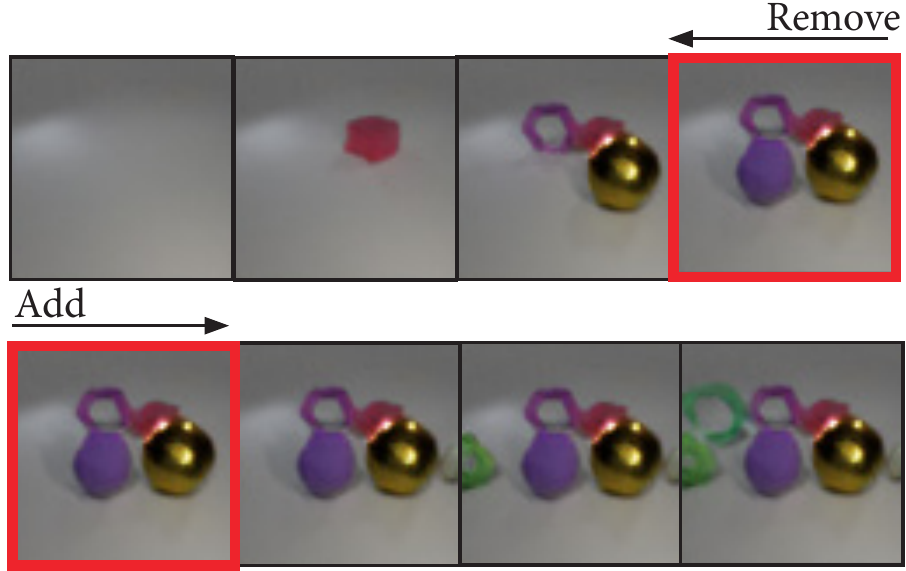}%
	\caption{\label{fig:add_remove}%
		Removing/adding objects.
		The red box shows the original scene.
	}
\end{wrapfigure}%
realistic shadows and highlights.
In Figure~\ref{fig:add_remove}, we use \name trained on \textsc{CLEVR4} and then remove (top) and add (bottom) more objects to the scene.
Note how \name generates realistic shadows and occlusion for scenes that the model has never seen before.

Secondly, we apply spatial manipulations that were not part of the similarity transform used during training, such as horizontal stretching, or slicing and combining different foreground objects.
Figure~\ref{fig:split_combine} shows that object features can be geometrically modified intuitively, without needing explicit 3D geometry or multi-view supervision during training.

\subsection{Comparison to 2D-based LR-GAN}

\begin{figure}
	\begin{minipage}{0.48\textwidth}%
		\includegraphics[width=\linewidth,trim=0 0 580 0,clip]{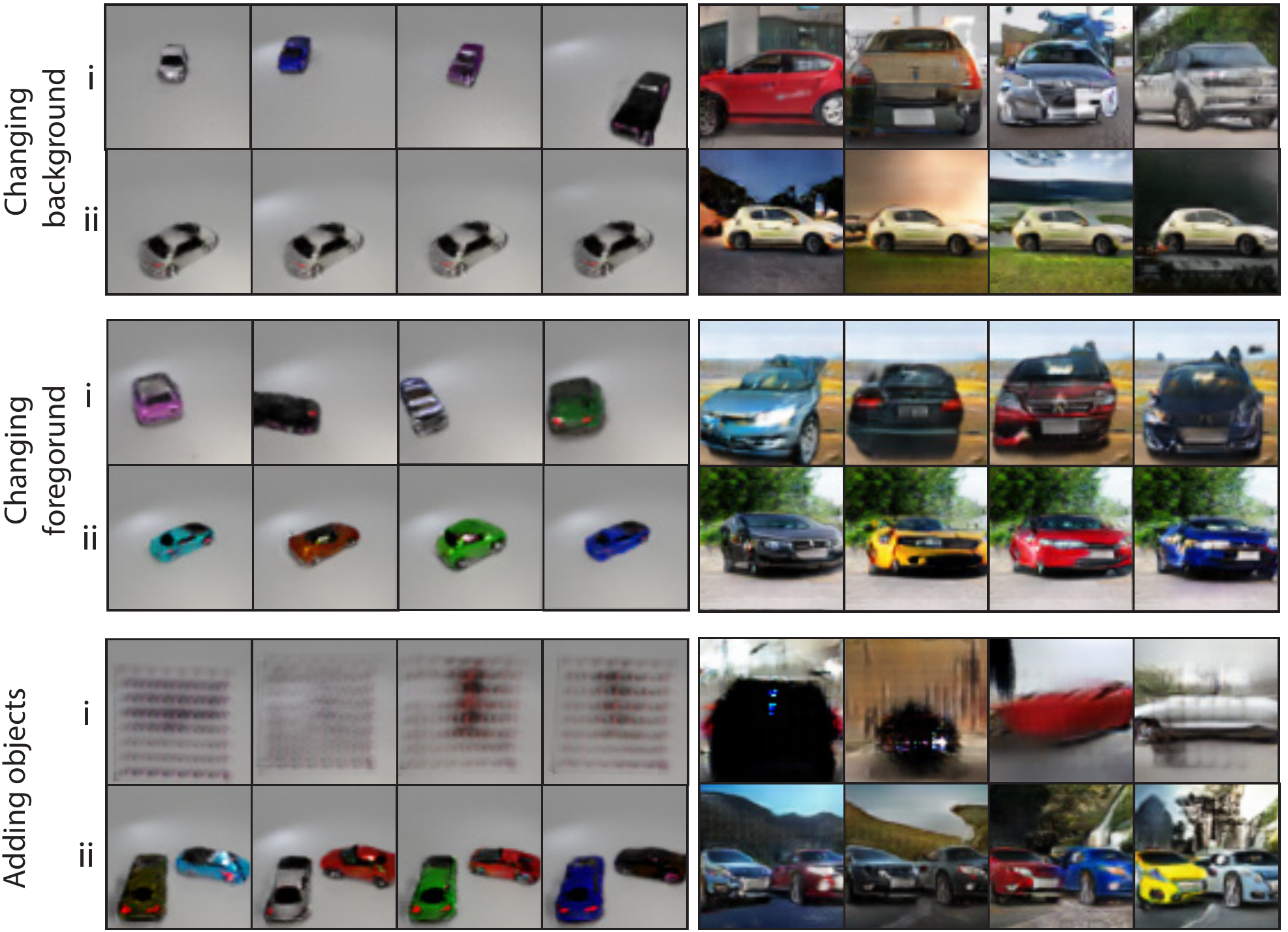}
		\captionof{figure}{\label{fig:compareLRGAN}
			Comparison between (i) LR-GAN \citep{yang_2017lr} and (ii) \name for \textsc{Synth-Car1} (left) and \textsc{Real-Car} (right).
		}
	\end{minipage}\hfill
	\begin{minipage}{0.48\textwidth}%
		\includegraphics[width=\linewidth,trim=580 0 0 0,clip]{Figures/compare_LRGAN_HoloGAN_combinedAblation.pdf}
		\captionof{figure}{\label{fig:Ablation_imbalanceView}
			Different manipulations applied to \name trained on
			(a) a dataset with imbalanced rotations, and
			(b) a balanced dataset.
		}
	\end{minipage}
\end{figure}

LR-GAN \citep{yang_2017lr} first generates a 2D background layer, and then generates and combines foreground layers with the generated background using alpha-compositing.
Both \name and LR-GAN show the importance of combining objects in a contextually relevant manner to generate visually realistic images (see Table~\ref{table:KID}).
However, LR-GAN does not offer explicit control over object location.
More importantly, LR-GAN learns an \emph{entangled} representation of the scene: sampling a different background noise vector also changes the foreground (Figure~\ref{fig:compareLRGAN}).
Finally, unlike \name, LR-GAN does not allow adding more foreground objects during test time.
This demonstrates the benefits of learning disentangled \emph{3D} object features compared to a 2D-based approach.

\subsection{Ablation study: Non-uniform pose distribution}
\label{sec:ablation}

For the natural \textsc{Real-Car} dataset, we observe that \name has difficulties learning the full 360° rotation of the car, even though fore- and background are disentangled well.
We hypothesise that this is due to the mismatch between the true (unknown) pose distribution of the car, and the uniform pose distribution we assume during training.
To test this, we create a synthetic dataset similar to \textsc{Synth-Car1} with a limited range of rotation, and train \name with a \emph{uniform} pose distribution. %
To generate the imbalanced rotation dataset, %
we sample the rotation uniformly from the front/left/back/right viewing directions $\pm 15$°.
In other words, the car is only seen from the front/left/back/right 30°, respectively, and there are four evenly spaced gaps of 60° that are never observed, for example views from the front-right.
With the imbalanced dataset, Figure~\ref{fig:Ablation_imbalanceView} (bottom) shows correct disentangling of foreground and background.
However, rotation of the car only produces images with (near-)frontal views (top), while depth translation results in cars that are randomly rotated sideways (middle).
We observe similar behaviour for the natural \textsc{Real-Car} dataset.
This suggests that learning object disentanglement and full 3D pose rotation might be two independent problems.
While assuming a uniform pose distribution already enables good object disentanglement, learning the pose distribution from the training data would likely improve the quality of 3D transforms.

In our supplemental material, we include comparisons to HoloGAN \cite{Nguyen-Phuoc_2019_ICCV} as well as additional ablation studies
on comparing different scene composer functions, %
using a perspective camera versus a weak-perspective camera, %
adopting the style discriminator for scenes with cluttered backgrounds, and %
training on images with an incorrect number of objects.

\section{Discussion and Future Work}

We introduced \name, an image generative model that learns 3D object-aware scene representations from unlabelled images.
We show that \name can learn a disentangled scene representation both in terms of objects and their properties, which allows geometric manipulations not observed during training.
Most excitingly, even when \name is trained with fewer or even single objects, additional 3D object features can be added to the scene features at test time to create novel scenes with multiple objects.
In addition to computer graphics applications, this opens up exciting possibilities, such as combining \name with models like BiGAN \cite{donahue2016bigan} or ALI \cite{dumoulin2016} to learn powerful object representations for scene understanding and reasoning.

Future work can adopt more powerful relational learning models \cite{NIPS2017_7082, NIPS2017_7181, Kipf2020Contrastive} to learn more complex object interactions such as inter-object shadowing or reflections. 
Currently, we assume prior knowledge of object category and the number of objects for training.
We also assume object poses are uniformly distributed and independent from each other.
Therefore, the ability to learn this information directly from training images would allow \name to be applied to more complex datasets with a varying number of objects and different object categories, such as COCO \cite{CoCo} or LSUN \cite{LSUN}.

\begin{ack}
We received support from the European Union's Horizon 2020 research and innovation programme under the Marie Skłodowska-Curie grant agreement No. 665992,
the EPSRC Centre for Doctoral Training in Digital Entertainment (EP/L016540/1),
RCUK grant CAMERA (EP/M023281/1),
an EPSRC-UKRI Innovation Fellowship (EP/S001050/1),
and an NVIDIA Corporation GPU Grant.
We received a gift from Adobe.
\end{ack}

\section*{Broader Impact}

\name is an image generative model that learns an object-oriented 3D scene representation directly from unlabelled 2D images.
Our approach is a new machine learning technique that makes it possible to generate unseen images from a noise vector, with unprecedented control over the identity and pose of multiple independent objects as well as the background.
In the long term, our approach could enable powerful tools for digital artists that facilitate artistic control over realistic procedurally generated digital content.
However, any tool can in principle be abused, for example by adding new, manipulating or removing existing objects or people from images.

At training time, our network performs a task somewhat akin to scene understanding, as our approach learns to disentangle between multiple objects and individual object properties (specifically their pose and identity).
At test time, our approach enables sampling new images with control over pose and identity for each object in the scene, but does not directly take any image input. However, it is possible to embed images into the latent space of generative models \cite{AbdalQW2019}.
A highly realistic generative image model and a good image fit would then make it possible to approximate the input image and, more importantly, to edit the individual objects in a pictured scene.
Similar to existing image editing software, this enables
the creation of image manipulations that could be used for ill-intended misinformation (\emph{fake news}), but also for a wide range of creative and other positive applications.
We expect the benefits of positive applications to clearly outweigh the potential downsides of malicious applications.

\bibliographystyle{plainnat}
\bibliography{ref_icml}

\clearpage\appendix

\section{Additional results}
\label{sec:moreresults}

\subsection{Comparison to \emph{entangled} 3D scene representation}

\begin{figure}[!htb]
	\centering
	\includegraphics[width=\linewidth]{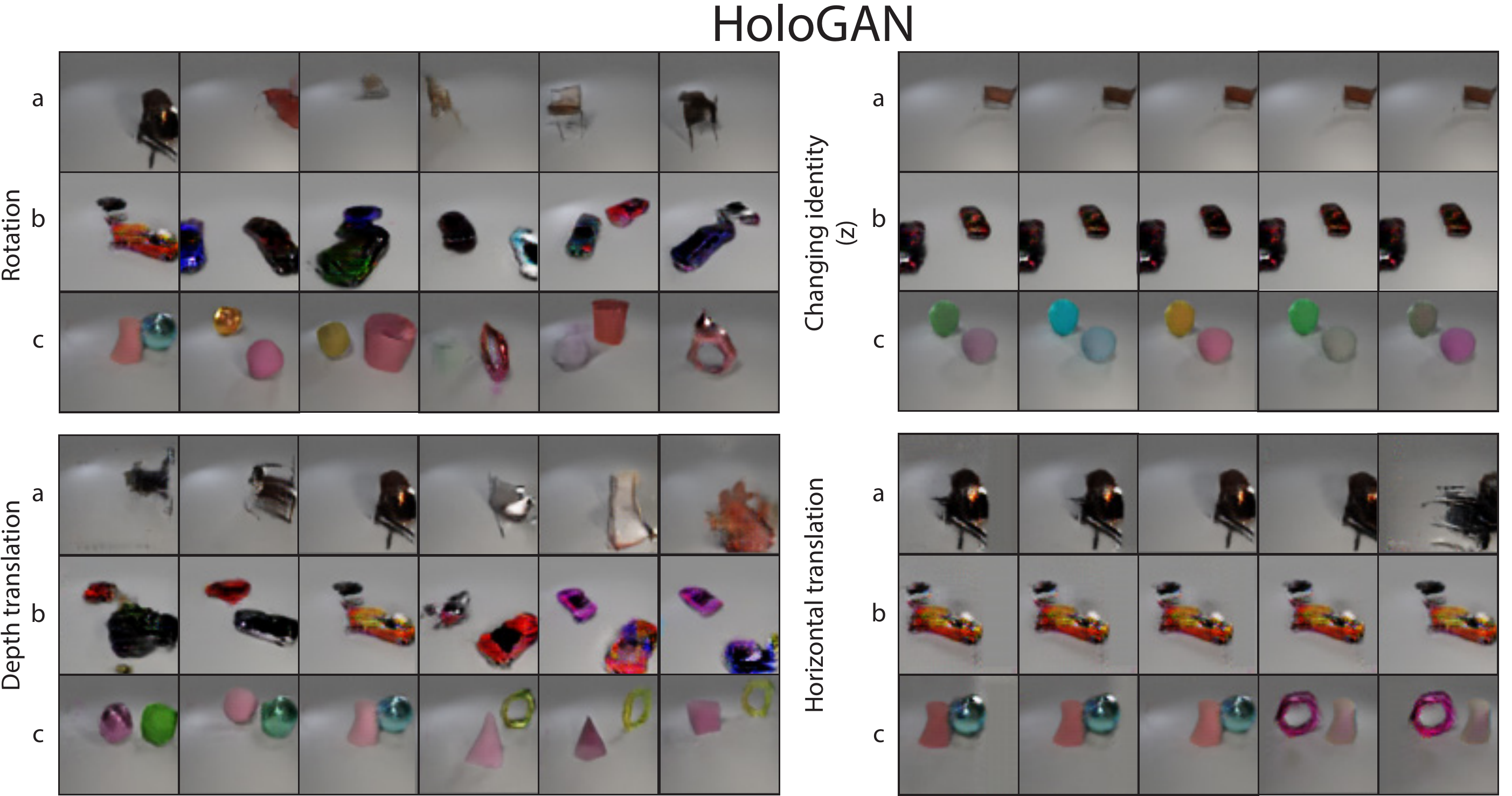}%
	\caption{\label{fig:baseline_compare_HoloGAN}%
		Samples from HoloGAN \cite{Nguyen-Phuoc_2019_ICCV} trained on the datasets (a) \textsc{Synth-Chair}1, (b) \textsc{Synth-Car2} and (c) \textsc{CLEVR2}.
		HoloGAN tends to associate each pose $\pose$ with a fixed object's identity, i.e., moving objects erroneously changes identity of both foreground and background (see top left, bottom left and right), while changing the noise vector $\mathbf{z}$ only changes a small part of the background (top right).
	}
\end{figure}

We compare \name with HoloGAN \cite{Nguyen-Phuoc_2019_ICCV}, which also learns deep 3D scene features but does not consider object disentanglement.
In particular, HoloGAN only considers one noise vector $\mathbf{z}$ for identity and one pose $\pose$ for the entire scene, and does not consider translation $\mathbf{t}$ as part of $\pose$.
While HoloGAN works well with object-centred scenes, it struggles with moving foreground objects.
Figure~\ref{fig:baseline_compare_HoloGAN} shows that HoloGAN tends to associate each pose $\pose$ with a fixed object's identity (i.e., moving objects erroneously changes identity of both foreground and background), while changing $\mathbf{z}$ only changes a small part of the background.
\name, on the other hand, can separate identity and pose for \emph{each} object, while being able to learn scene-level effects such as lighting and shadows.

\subsection{Increasing the number of foreground objects}

\begin{figure}[b!]
	\centering
	\includegraphics[width=0.70\linewidth]{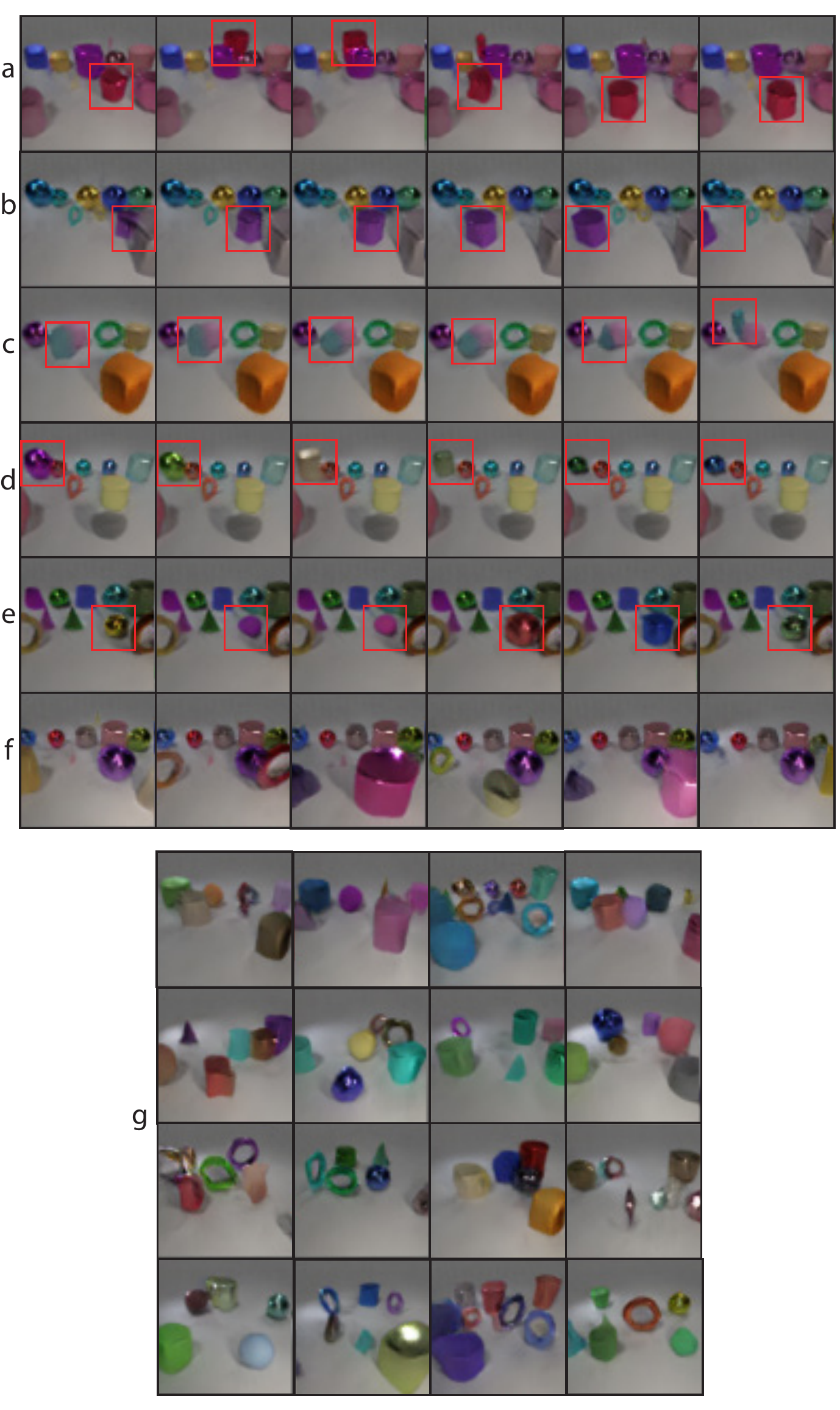}%
	\caption{\label{fig:failureMode}%
		Qualitative results of \name trained on \textsc{CLEVR6}.
		(a) Rotating a foreground object,
		(b) Horizontal translation,
		(c) Depth translation,
		(d) Changing foreground object 1,
		(e) Changing foreground object 3,
		(f) Changing the background object,
		and (g) Random samples.}
\end{figure}

To test the capability of our method, we train \name on CLEVR6 (6 foreground objects).
As shown in Figure~\ref{fig:failureMode}, \name is still capable of generating and manipulating 3D object features, although the background generator now also produces foreground objects (Figure~\ref{fig:failureMode}f).
Moreover, rotating individual object leads to changes in object's depth (Figure~\ref{fig:failureMode}a).

Interestingly, we notice that \name now generates images with more or less than 6 objects, despite being trained with images that contain exactly 6 objects (Figure~\ref{fig:failureMode}g).
We hypothesise that BlockGAN's failure in this case is due to our assumption that the poses of all objects are independent from each other (during training, we randomly sample the pose $\theta$ for each object).
This is not true in the physical world (also in the CLEVR dataset) where objects do not intersect.
The more objects there are in the scene, the stronger the interdependence between objects' poses becomes.
Therefore, for future work, we hope to adopt more powerful relation learning structures to learn objects' pose directly from training images.
Another interesting direction is to design object-aware discriminators, which are capable of recognising fake images when the generators produce samples with more objects than the training images.

\section{Additional ablation studies}
\label{sec:moreablations}

\subsection{Learning without the perspective camera}

\begin{figure}[!htb]
 	\centering
	\includegraphics[width=0.7\linewidth]{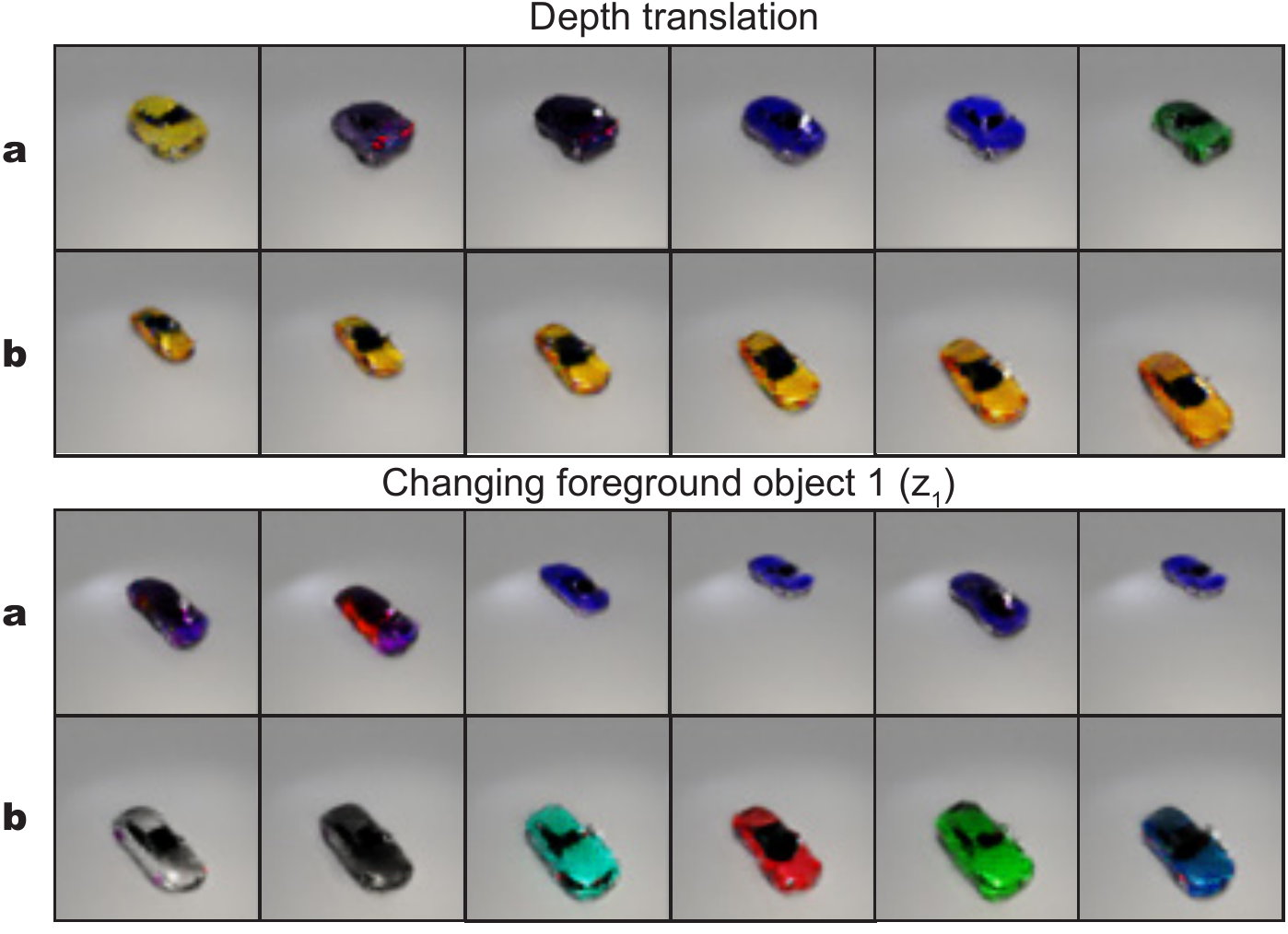}%
	\caption{\label{fig:Ablation_noSkew}%
		The effect of modelling the perspective camera explicitly (b) compared to using a weak-perspective camera (a).
		Note that with the weak-perspective camera (a), translation along the depth dimension (top) leads to identity changes without any translation in depth, while changing the noise vector $\mathbf{z}_1$ (bottom) changes both depth translation and, to a lesser extent, the object identity.
		Using a perspective camera correctly disentangles position and identity (b).
	}
\end{figure}

Here we show the advantage of implementing the perspective camera explicitly, compared to using a weak-perspective projection like HoloGAN \citep{Nguyen-Phuoc_2019_ICCV}.
Since a perspective camera directly affects foreshortening, it provides strong cues for \name to solve the scale/depth ambiguity.
This is especially important for \name to learn to project and reason over occlusion by concatenating the depth and channel dimension, followed by an MLP.
Since the MLP is flexible, \name trained \emph{without} a perspective camera, therefore, tends to learn to associate an object's identity with scale and depth, while changing depth only changes the object's appearance (see Figure~\ref{fig:Ablation_noSkew}).

\subsection{Scene composer function}

We consider and compare three scene composer functions: (i)~element-wise summation, (ii)~element-wise maximum, and (iii)~an MLP (multi-layer perceptron).
We train \name with each function and compare their performance in terms of visual quality (KID score)
in Table~\ref{table:KID2}.
While all three functions can successfully combine objects into a scene, the element-wise maximum performs best and easily generalises to multiple objects.
Therefore, we use the element-wise maximum for \name.

\begin{table}[h!]
	\caption{\label{table:KID2}%
		KID estimates for different scene composer functions.
	}
	\centering
	{%
		\newcommand{\size}{{\footnotesize 64$\times$64}}%
		\begin{tabular}{lccc}
			\textbf{Method} &     \textsc{Synth-Car1} (\size)      &     \textsc{Synth-Chair1} (\size)     &  \\ \midrule
			Sum             &          0.040 $\pm$ 0.002           &           0.038 $\pm$ 0.001           &  \\
			MLP             &          0.044 $\pm$ 0.001           &           0.033 $\pm$ 0.001           &  \\
			Max             &      \textbf{0.039 $\pm$ 0.002}      &     \textbf{0.031 $\pm$ 0.001} %
	\end{tabular}}
\end{table}

\subsection{Learning without the style discriminator}

\begin{figure}[h]
 	\centering
	\includegraphics[width=0.7\linewidth]{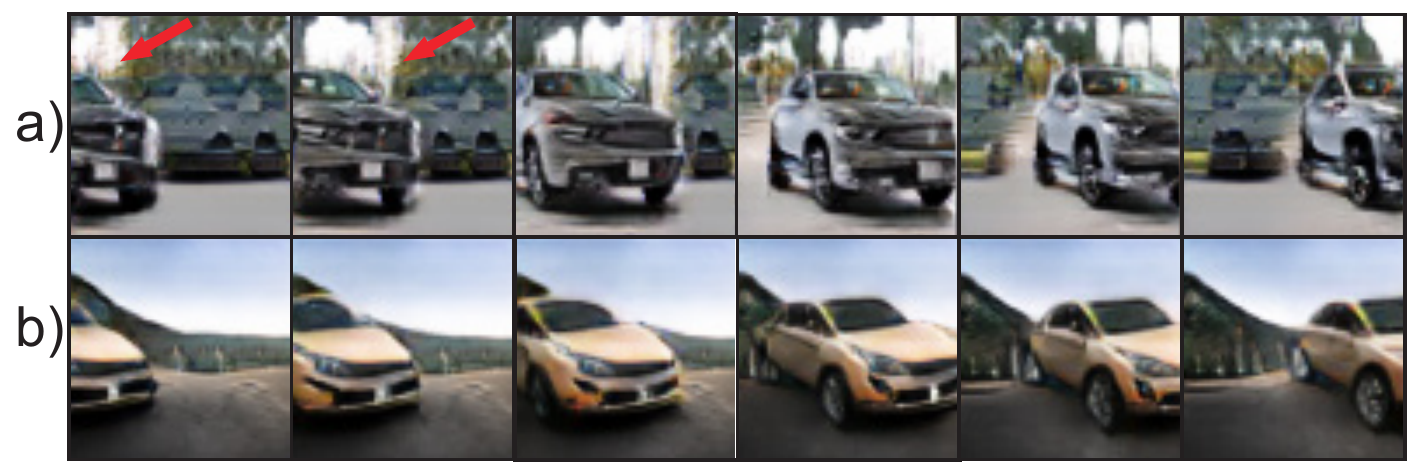}%
	\caption{\label{fig:Ablation_StyleDisc}%
		With a standard discriminator (a), a part of the background appearance is baked into the foreground object (see red arrows).
		Adding the style discriminator (b) cleanly separates the car from the background.
	}
\end{figure}

When \name is trained with a standard discriminator on datasets with a cluttered background, such as the \textsc{Real-Car} dataset, the foreground object features tend to include part of the background object.
This creates visual artefacts when objects move in the scene (indicated by red arrows in Figure~\ref{fig:Ablation_StyleDisc}a).
We hypothesise that these artefacts should be picked up by the discriminator since generated images should look unrealistic.
Therefore, we add more powerful style discriminators \citep{Nguyen-Phuoc_2019_ICCV} to the original discriminator at different layers (see Section~\ref{sec:supploss} for details).
Figure~\ref{fig:Ablation_StyleDisc}b shows that the generator is indeed discouraged from adding background information to the foreground object features, leading to cleaner results.

\subsection{Incorrect number of objects}

\begin{figure}[h]
	\centering
	\includegraphics[width=0.49\linewidth,trim=0 162 0 0,clip]{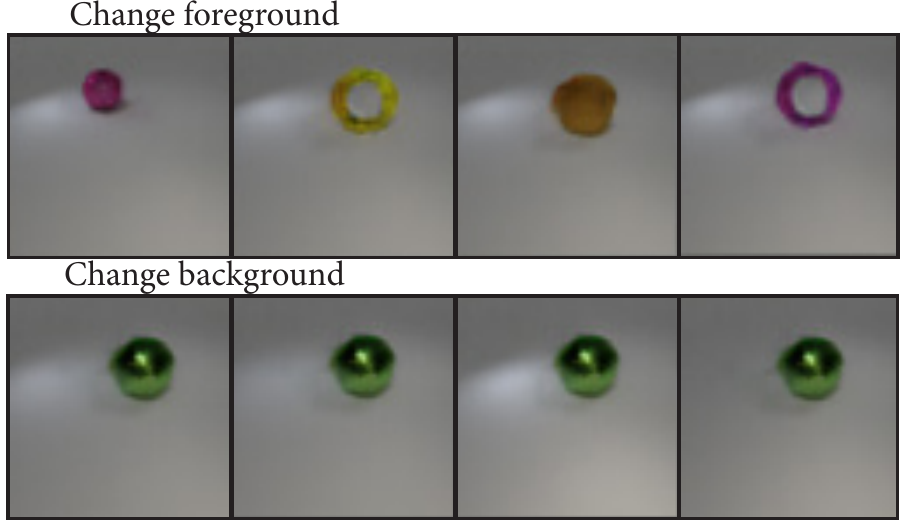}\hfill%
	\includegraphics[width=0.49\linewidth,trim=0 0 0 162,clip]{Figures/rebuttal_wrongNumberOfObjects.pdf}
	\caption{\label{fig:Ablation_NumObjects}%
		Results for BlockGAN trained with 2 foreground (FG) object generators when trained on 1 or 3 foreground objects.
		1 object (left): Changing either FG object changes the object's appearance and pose; changing the background works as expected.
		3 objects (right): Changing one FG object changes one object as expected; changing the background changes one FG object and the background.
	}
\end{figure}

We next investigate the performance of BlockGAN when the training data contains fewer or more objects than expected.
In Figure~\ref{fig:Ablation_NumObjects}, we show BlockGAN configured with 2 foreground object generators when trained with images containing 1 or 3 foreground objects.
If only a single object is present (Figure~\ref{fig:Ablation_NumObjects}, left), changing either of the two foreground generators changes the object's appearance and pose (top), while changing the background works as expected (bottom).
If there are three objects present (Figure~\ref{fig:Ablation_NumObjects}, right), changing one foreground generator changes one object as expected (top), while changing the background generator simultaneously changes one foreground object and the background (bottom).

\section{Comparison to other methods}
\label{sec:comparisons}

\begin{figure}
 	\centering
	\includegraphics[width=\linewidth]{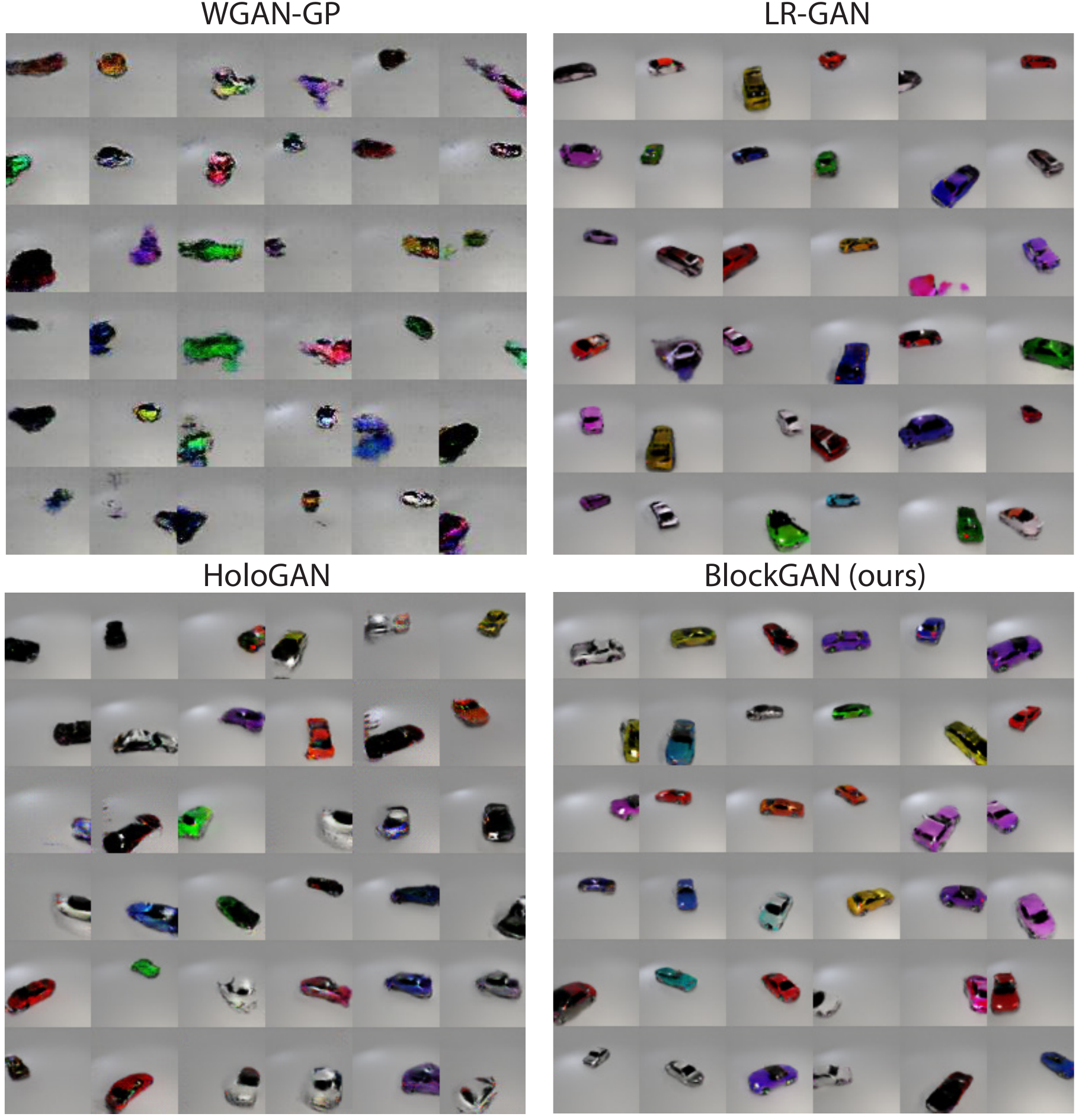}%
	\caption{\label{fig:baseline_compare_synthCar}%
		Samples from WGAN-GP, LR-GAN, HoloGAN and our BlockGAN trained on \textsc{Synth-Car1}.}
\end{figure}

\begin{figure}
 	\centering
	\includegraphics[width=\linewidth]{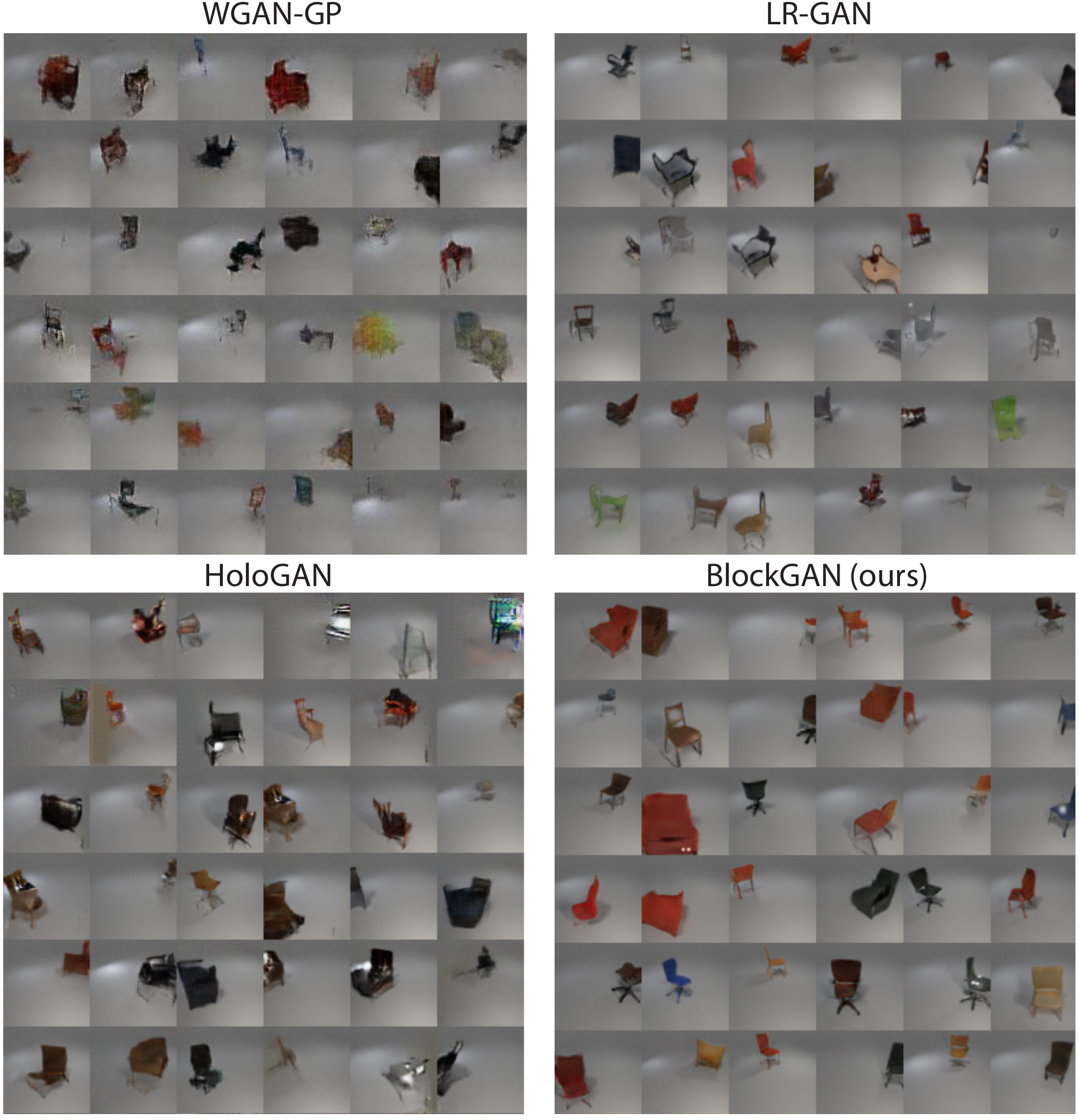}%
	\caption{\label{fig:baseline_compare_SynthChair}%
		Samples from WGAN-GP, LR-GAN, HoloGAN and our BlockGAN trained on \textsc{Synth-Chair1}.}
\end{figure}

\begin{figure}
 	\centering
	\includegraphics[width=\linewidth]{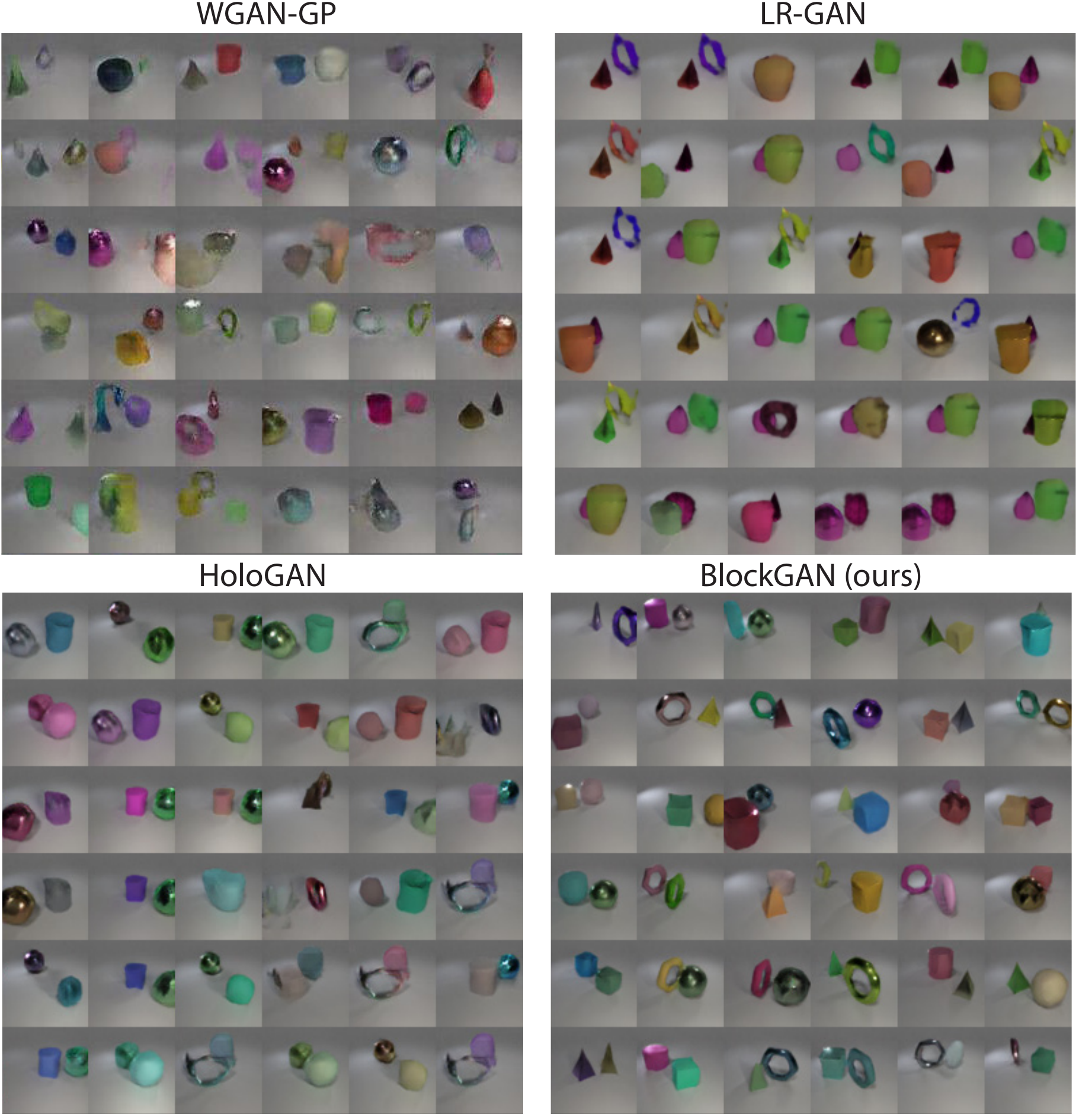}%
	\caption{\label{fig:baseline_compare_CLEVR2}%
		Samples from WGAN-GP, LR-GAN, HoloGAN and our BlockGAN trained on \textsc{CLEVR2}.}
\end{figure}

\begin{figure}
 	\centering
	\includegraphics[width=\linewidth]{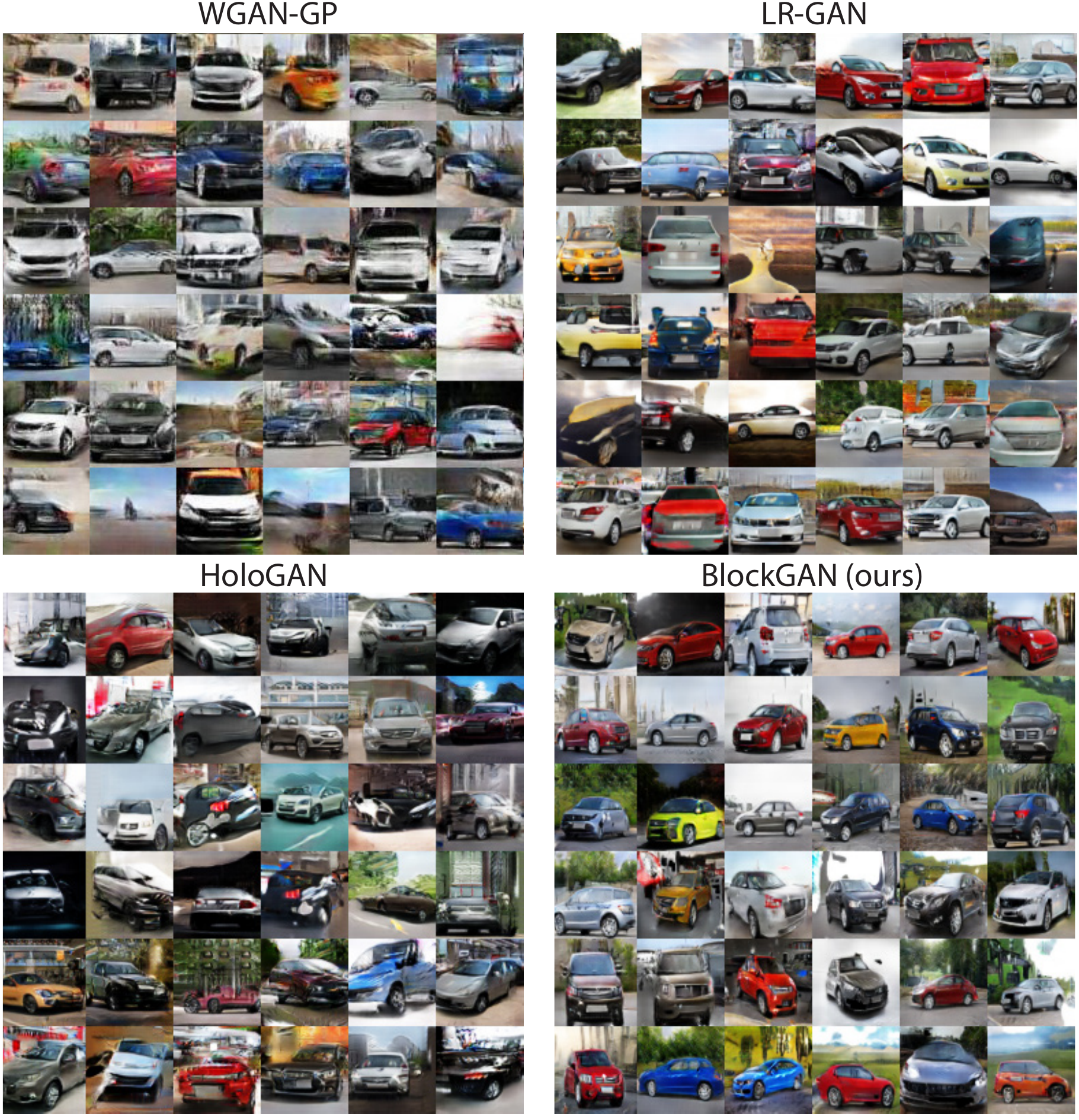}%
	\caption{\label{fig:baseline_compare_RealCars}%
		Samples from WGAN-GP, LR-GAN, HoloGAN and our BlockGAN trained on \textsc{Real-Cars}.}
\end{figure}

In Figure~\ref{fig:baseline_compare_synthCar}, \ref{fig:baseline_compare_SynthChair}, \ref{fig:baseline_compare_CLEVR2} and \ref{fig:baseline_compare_RealCars}, we show generated samples by a vanilla GAN (WGAN-GP \citep{NIPS2017_7159}), 2D object-aware LR-GAN \citep{yang_2017lr}, 3D-aware HoloGAN \citep{Nguyen-Phuoc_2019_ICCV} and our \name.
Compared to other models, \name produces samples with competitive or better quality, \emph{and} offers explicit control over the poses of objects in the generated images.
Notice that although LR-GAN is designed to handle foreground and background objects explicitly, for \textsc{CLEVR2} with two foreground objects, this method struggles and tends to always place one foreground object at the image centre (see Figure~\ref{fig:baseline_compare_CLEVR2}).

\paragraph{Implementation details} 

For WGAN-GP, we use a publicly available implementation\footnote{\href{https://github.com/LynnHo/DCGAN-LSGAN-WGAN-WGAN-GP-Tensorflow}{https://github.com/LynnHo/DCGAN-LSGAN-WGAN-WGAN-GP-Tensorflow}}\!\!.
For LR-GAN and HoloGAN, we use the code provided by the authors.
We conduct hyperparameter search for these models, and report best results for each method.
Note that for HoloGAN, we modify the 3D transformation to add translation during training, since this method assumes that foreground objects are at the image centre.

\clearpage

\section{Loss function and style discriminator}
\label{sec:supploss}

For datasets with cluttered backgrounds like the natural \textsc{Real-Car} dataset, we adopt \emph{style discriminators} in addition to the normal image discriminator (see the benefit in Figure~\ref{fig:Ablation_StyleDisc}).
Style discriminators perform the same real/fake classification task as the standard image discriminator, but at the feature level across different layers.
In particular, style discriminators classify the mean $\boldsymbol\mu$ and standard deviation $\boldsymbol\sigma$ of the features $\mathbf{\Phi}_l$ at different levels $l$ (which are believed to describe the image ``style'').
The mean $\boldsymbol\mu(\mathbf{\Phi}_l(\mathbf{x}))$ and variance $\boldsymbol\sigma(\mathbf{\Phi}_l(\mathbf{x}))$ of the features $\mathbf{\Phi}_l(\mathbf{x})$ are computed across batch and spatial dimensions independently using:
\begin{align}
\boldsymbol\mu(\mathbf{\Phi}_l(\mathbf{x})) &= \frac{1}{N \times H \times W} \sum_{n=1}^{N} \sum_{h=1}^{H} \sum_{w=1}^{W} \mathbf{\Phi}_l(\mathbf{x})_{nhw} \text{,} \\
\boldsymbol\sigma(\mathbf{\Phi}_l(\mathbf{x})) &= \sqrt{\frac{1}{N \times H \times W} \sum_{n=1}^{N} \sum_{h=1}^{H} \sum_{w=1}^{W}
	\big(\mathbf{\Phi}_l(\mathbf{x})_{nhw} - \boldsymbol\mu(\mathbf{\Phi}_l(\mathbf{x}))\big)^2 + \epsilon} \text{.}
\end{align}

The style discriminators are implemented as MLPs with sigmoid activation functions for binary classification.
A style discriminator at layer $l$ is written as
\begin{align}
L_\text{style}^l(\text{G}) &= \mathbb{E}_\mathbf{z, \theta} [-\log \; {\text{D}_l(\text{G}(\mathbf{z, \theta}))}] \text{.}
\end{align}
The total loss therefore can be written as
\begin{align}
L_\text{total}(\text{G}) &= L_\text{GAN}(\text{G})
+ \lambda_\text{s} \!\cdot\! \sum_{l} L_\text{style}^l(\text{G})
\text{.}
\end{align}
We set $\lambda_\text{s} = 1$ for all natural datasets and $\lambda_\text{s} = 0$ for synthetic datasets.

\section{Datasets}
\label{sec:dataset}

\begin{table}[b!]
	\caption{\label{tab:datasets}%
		Datasets used in our paper ($n$ = number of foreground objects).
		‘Azimuth’ describes object rotation about the up-axis.
		‘Elevation’ refers to the camera's elevation above ground.
		‘Scaling’ is the scale factor applied to foreground objects.
		‘Horiz. transl.’ and ‘Depth transl.’ are horizontal/depth translation of objects relative to the global origin.
		Ranges represent uniform random distributions.
	}\vspace{0.5em}
	\resizebox{\linewidth}{!}{%
		\renewcommand*{\arraystretch}{1.2}%
		\begin{tabular}{lrrrccc}
			\toprule
			\textbf{Name}                   & \textbf{\# Images} & \textbf{Azimuth} & \textbf{Elevation} & \textbf{Scaling} & \textbf{Horiz. transl.} & \textbf{Depth transl.} \\ \midrule
			\textsc{Synth-Car$n$}             &             80,000 &        0° – 359° &                45° &        0.5 – 0.6 &          –5 – 5         &         –5 – 5         \\ \gline
			\textsc{Synth-Chair$n$}            &            100,000 &        0° – 359° &                45° &        0.5 – 0.6 &          –5 – 5         &         –5 – 5         \\ \gline
			\textsc{CLEVR$n$} \citep{johnson2017clevr} &            100,000 &        0° – 359° &          45° &        0.5 – 0.6 &          –4 – 4         &         –4 – 4      \\ \gline
			\textsc{Real-Cars} \citep{Yangg2015} &            139,714 &        0° – 359° &           0° – 35° &        0.5 – 0.8 &          –3 – 4         &         –5 – 6      \\\bottomrule
		\end{tabular}%
	}
\end{table}

\begin{figure}
 	\centering
	\includegraphics[width=\linewidth]{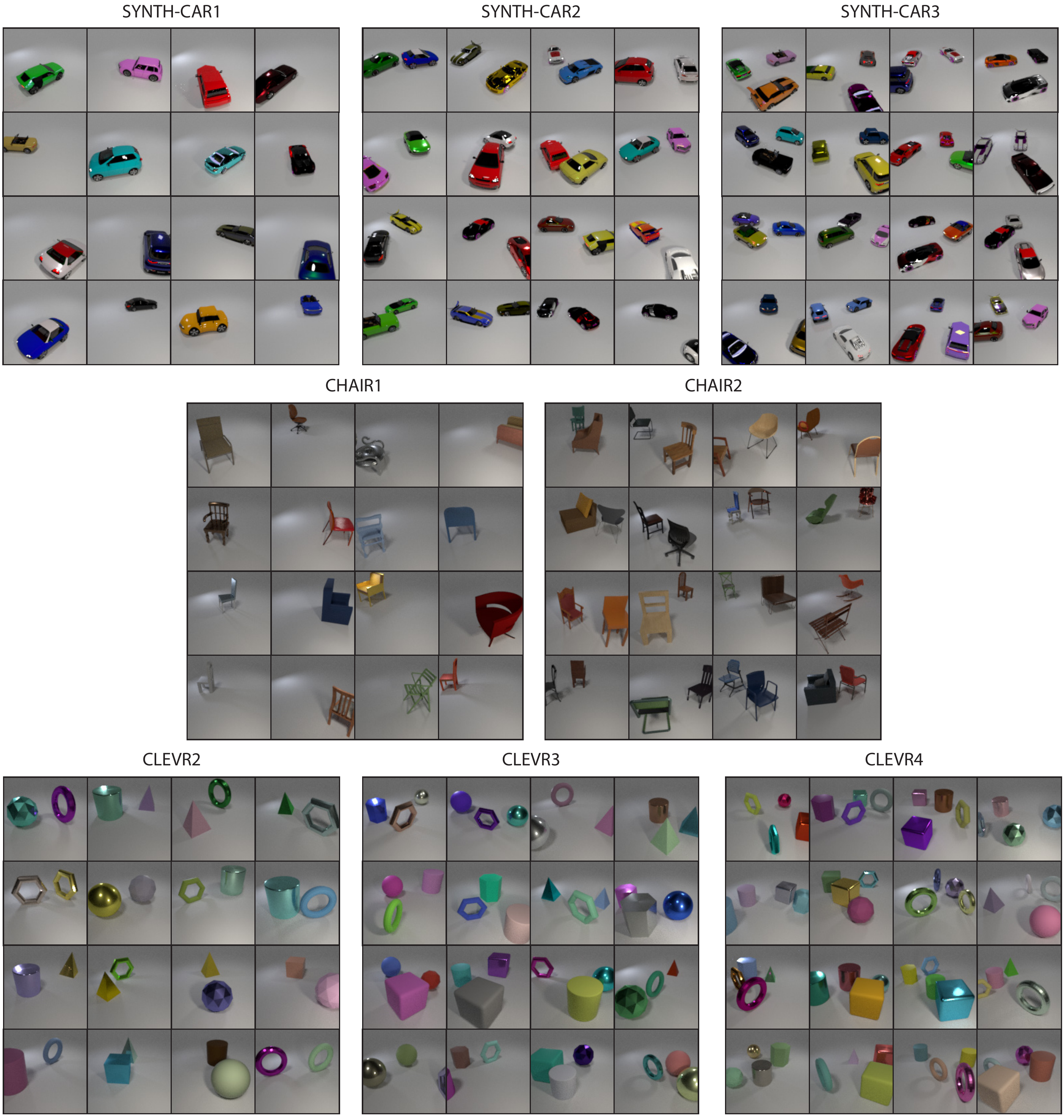}%
	\caption{\label{fig:dataset_samples}%
		Samples from the synthetic datasets.
	}
\end{figure}

We modify the CLEVR dataset \citep{johnson2017clevr} to add a larger variety of colours and primitive shapes.
Additionally, we use the scene setups provided by CLEVR to render the remaining synthetic datasets (\textsc{Synth-Car$n$} and \textsc{Synth-Chair$n$}, with $n$ foreground objects each).
These include a fixed, grey background, a virtual camera with fixed parameters but random location jittering, and random lighting.
We also use the render script from CLEVR to randomly place foreground objects into the scene and render them.
We render all image at resolution 128 $\times$ 128, and bi-linearly downsample them to 64 $\times$ 64 for training.
For the natural \textsc{Car} dataset, each image is first scaled such that the smaller side is 64, then it is cropped to produce a 64$\times$64\, pixel crop.
During training, we randomly move the 64$\times$64 cropping window before cropping the image.
Figure~\ref{fig:dataset_samples} includes samples from our generated datasets, and
Table~\ref{tab:datasets} lists the range of pose parameters used for each dataset during training.

Link for 3D textured chair models:\\
\url{https://keunhong.com/publications/photoshape/}

Link for CLEVR:\\
\url{https://github.com/facebookresearch/clevr-dataset-gen}

Link for natural \textsc{Car} dataset:\\
\url{http://mmlab.ie.cuhk.edu.hk/datasets/comp_cars/}

\section{Implementation}
\label{sec:implementation}

\subsection{Training details}

\paragraph{Virtual camera model.}

We assume a virtual camera with a focal length of 35\,mm and a sensor size of 32\,mm (\citetalias{blender}'s default values), which corresponds to an angle of view of $2\arctan \frac{32\,\text{mm}}{2 \times  35\,\text{mm}} \!=\! 49.1$\,degrees (we use the same setup for natural images).

\paragraph{Sampling.}

We initialise all weights using $\mathcal{N}(0, 0.2)$ and biases as $0$.
For \textsc{CLEVR$n$}, we use noise vector dimensions of $\abs{\mathbf{z}_0} = 20$ for the background, and $\abs{\mathbf{z}_i} = 60$ (for $i=1,\ldots,n$) for the foreground objects, to account for their relative visual complexity.
Similarly, for \textsc{Synth-Car$n$} and \textsc{Synth-Chair$n$}, we use $\abs{\mathbf{z}_0} = 30$ and $\abs{\mathbf{z}_i} = 90$ (for $i = 1, \ldots, n$), to account for their relative visual complexity.
For the natural \textsc{Real-Car} dataset, we use $\abs{\mathbf{z}_0} = 100$ and $\abs{\mathbf{z}_1} = 200$.
Note that we only feed $\mathbf{z}$ to the 3D features of each object, and not to the 3D scene features and 2D features.
Table~\ref{tab:datasets} provides the ranges we use for sampling the pose $\boldsymbol\theta_i$ of foreground objects during training.

\paragraph{Training.}

We train \name using the Adam optimiser \citep{adam}, with $\beta_1 = 0.5$ and $\beta_2 = 0.999$.
We use the same learning rate for both the discriminator and the generator.
Empirically, we find that updating the generator twice for every update of the discriminator achieves images with the best visual fidelity.
We use a learning rate of 0.0001 for all synthetic datasets.
For the natural \textsc{Cars} dataset, we use a learning rate of 0.00005.
We train all datasets with a batch size of 64 for 50 epochs.
Training takes 1.5 days for the synthetic datasets and 3 days for the natural \textsc{Real-Cars} dataset.

\paragraph{Infrastructure.}

All models were trained using a single GeForce RTX 2080 GPU.

\subsection{Network architecture}

We describe the network architecture for the \name foreground object generator in Table~\ref{tab:fg}, the \name background generator in Table~\ref{tab:bg}, and the overall \name generator in Tables~\ref{tab:synthetic-generator} and \ref{tab:natural-generator} for synthetic and real datasets, respectively.
Note that we use ReLU for the synthetic datasets and LReLU for the natural \textsc{Car} dataset after the AdaIN layer.
The discriminator is described in Table~\ref{tab:discriminator}.

In terms of the notation in Section~3 of the main paper, object features have dimensions $H_o \times W_o \times D_o \times C_o = 16 \times 16 \times 16 \times 64$,
scene features have the same dimensions $H_s \times W_s \times D_s \times C_s = 16 \times 16 \times 16 \times 64$, and
camera features have dimensions $H_c \times W_c = 16 \times 16$ (before up-convolutions to $64 \times 64$) with $C_c = 64$ channels for synthetic datasets and $C_c = 256$ channels for natural image datasets.

As GANs empirically tend to perform better on category-specific datasets, we decided to start with this assumption.
A promising future direction is to adopt a shared rendering layer for objects generated by different category-specific generators, similar to \citet{AlievSKUL2020}.

\begin{table}[h]
	\vspace{2em}
	\caption{\label{tab:fg}%
		Network architecture of the \name foreground (FG) object generator.
	}%
	\centering
	{%
		\renewcommand*{\arraystretch}{1.2}%
		\begin{tabular}{lcccc}
			\toprule
			\textbf{Layer type}    & \textbf{Kernel size}  & \textbf{Stride} & \textbf{Normalisation} &     \textbf{Output dimension}      \\ \midrule
			Learnt constant tensor &           —           &        —        &         AdaIN          &  $4 \times 4 \times 4 \times 512$  \\ \gline
			UpConv                 & $3 \times 3 \times 3$ &        2        &         AdaIN          &  $8 \times 8 \times 8 \times 128$  \\ \gline
			UpConv                 & $3 \times 3 \times 3$ &        2        &         AdaIN          & $16 \times 16 \times 16 \times 64$ \\ \gline
			3D transformation      &           —           &        —        &           —            & $16 \times 16 \times 16 \times 64$ \\ \bottomrule
	\end{tabular}}
	
	\vspace{3em}
	
	\caption{\label{tab:bg}%
		Network architecture of the \name background (BG) object generator.
	}%
	\centering
	{%
		\renewcommand*{\arraystretch}{1.2}%
		\begin{tabular}{lcccc}
			\toprule
			\textbf{Layer type}    & \textbf{Kernel size}  & \textbf{Stride} & \textbf{Normalisation} &     \textbf{Output dimension}      \\ \midrule
			Learnt constant tensor &           —           &        —        &         AdaIN          &  $4 \times 4 \times 4 \times 256$  \\ \gline
			UpConv                 & $3 \times 3 \times 3$ &        2        &         AdaIN          &  $8 \times 8 \times 8 \times 128$  \\ \gline
			UpConv                 & $3 \times 3 \times 3$ &        2        &         AdaIN          & $16 \times 16 \times 16 \times 64$ \\ \gline
			3D transformation      &           —           &        —        &           —            & $16 \times 16 \times 16 \times 64$ \\ \bottomrule
	\end{tabular}}
\end{table}

\begin{table}
	\caption{\label{tab:synthetic-generator}%
		Network architecture of the \name generator for all synthetic datasets.
	}%
	\centering
	\resizebox{\linewidth}{!}
	{%
		\renewcommand*{\arraystretch}{1.2}%
		\begin{tabular}{lccccc}
			\toprule
			\textbf{Layer type}               & \textbf{Kernel size} & \textbf{Stride} & \textbf{Activation} & \textbf{Norm.} &      \textbf{Output dimension}      \\ \midrule
			$n\times$FG generator (Table~\ref{tab:fg}) &          —           &        —        &        ReLU         &       —        & $16 \times 16 \times 16 \times 64$  \\ \gline
			BG generator (Table~\ref{tab:bg}) &          —           &        —        &        ReLU         &       —        & $16 \times 16 \times 16 \times 64$  \\ \gline
			Element-wise maximum              &          —           &        —        &          —          &       —        & $16 \times 16 \times 16 \times 64$  \\ \gline
			Concatenate                       &          —           &        —        &          —          &       —        & $16 \times 16 \times (16 \cdot 64)$ \\ \gline
			Conv                              &     $1 \times 1$     &        1        &        ReLU         &       —        &      $16 \times 16 \times 64$       \\ \gline
			UpConv                            &     $4 \times 4$     &        2        &        ReLU         &     AdaIN      &      $32 \times 32 \times 64$       \\ \gline
			UpConv                            &     $4 \times 4$     &        2        &        ReLU         &     AdaIN      &      $64 \times 64 \times 64$       \\ \gline
			UpConv                            &     $4 \times 4$     &        1        &        ReLU         &     AdaIN      &       $64 \times 64 \times 3$       \\ \bottomrule
	\end{tabular}}
	
	\vspace{3em}
	
	\caption{\label{tab:natural-generator}%
		Network architecture of the \name generator for the \textsc{Real-Cars} dataset.
		Differences to the synthetic foreground object generator in Table~\ref{tab:synthetic-generator} are \diff{highlighted in blue}.
	}%
	\resizebox{\linewidth}{!}{%
		\renewcommand*{\arraystretch}{1.2}%
		\begin{tabular}{lccccc}
			\toprule
			\textbf{Layer type}               & \textbf{Kernel size} & \textbf{Stride} & \textbf{Activation} & \textbf{Normal.} &      \textbf{Output dimension}      \\ \midrule
			FG generator (Table~\ref{tab:fg}) &          —           &        —        &     \diff{LReLU}    &        —         & $16 \times 16 \times 16 \times 64$  \\ \gline
			BG generator (Table~\ref{tab:bg}) &          —           &        —        &     \diff{LReLU}    &        —         & $16 \times 16 \times 16 \times 64$  \\ \gline
			Element-wise maximum              &          —           &        —        &          —          &        —         & $16 \times 16 \times 16 \times 64$  \\ \gline
			Concatenate                       &          —           &        —        &          —          &        —         & $16 \times 16 \times (16 \cdot 64)$ \\ \gline
			Conv                              &     $1 \times 1$     &        1        &     \diff{LReLU}    &        —         &  $16 \times 16 \times \diff{256}$   \\ \gline
			UpConv                            &     $4 \times 4$     &        2        &     \diff{LReLU}    &      AdaIN       &  $32 \times 32 \times \diff{128}$   \\ \gline
			UpConv                            &     $4 \times 4$     &        2        &     \diff{LReLU}    &      AdaIN       &      $64 \times 64 \times 64$       \\ \gline
			UpConv                            &     $4 \times 4$     &        1        &     \diff{LReLU}    &      AdaIN       &       $64 \times 64 \times 3$       \\ \bottomrule
	\end{tabular}}
	
	\vspace{3em}
	
	\caption{\label{tab:discriminator}%
		Network architecture of the \name discriminator for both synthetic and real datasets.
	}%
	\centering
	{%
		\renewcommand*{\arraystretch}{1.2}%
		\begin{tabular}{lccccc}
			\toprule
			\textbf{Layer type} & \textbf{Kernel size} & \textbf{Stride} & \textbf{Activation} & \textbf{Normalisation} & \textbf{Output dimension} \\ \midrule
			Conv                &     $5 \times 5$     &        2        &        LReLU        &      IN/Spectral       & $32 \times 32 \times 64$  \\ \gline
			Conv                &     $5 \times 5$     &        2        &        LReLU        &      IN/Spectral       & $16 \times 16 \times 128$ \\ \gline
			Conv                &     $5 \times 5$     &        2        &        LReLU        &      IN/Spectral       &  $8 \times 8 \times 256$  \\ \gline
			Conv                &     $5 \times 5$     &        2        &        LReLU        &      IN/Spectral       &  $4 \times 4 \times 512$  \\ \gline
			Fully connected     &          —           &        —        &       Sigmoid       &     None/Spectral      &            $1$            \\ \bottomrule
	\end{tabular}}
\end{table}

\end{document}